\documentclass[10pt,twocolumn,letterpaper]{article}

\usepackage[pagenumbers]{cvpr} % To force page numbers, e.g. for an arXiv version
\usepackage{array,tabularx,booktabs,multirow,pifont}
\newcommand{\cmark}{\textcolor{ForestGreen}{\ding{51}}} % ✓
\newcommand{\xmark}{\textcolor{Red}{\ding{55}}} % ✗
\usepackage[table]{xcolor} % if not already loaded
\definecolor{rowhl}{RGB}{235,245,255} % very light blue
\definecolor{badresult}{RGB}{180,30,30}
\usepackage{multirow}
\usepackage[normalem]{ulem}
\useunder{\uline}{\ul}{}
\definecolor{vlgray}{RGB}{242,242,242} % very light gray

\usepackage[margin=1in]{geometry}
\usepackage{booktabs}
\usepackage{siunitx}
\usepackage{makecell}
\usepackage{graphicx}
\usepackage{xspace}
\usepackage{xfrac}
\newcommand{\ourmethod}{XL-VLA\xspace}

\definecolor{gqorange}{HTML}{EE4431}

\newcommand{\impr}[1]{\textcolor{red}{{\scriptsize\selectfont +#1}}}
\definecolor{cvprblue}{rgb}{0.21,0.49,0.74}
\usepackage[pagebackref,breaklinks,colorlinks,allcolors=cvprblue]{hyperref}

\def\paperID{2752} % *** Enter the Paper ID here
\def\confName{CVPR}
\def\confYear{2026}

\title{Cross-Hand Latent Representation for Vision-Language-Action Models}

\author{
Guangqi Jiang$^{1*}$\quad Yutong Liang$^{1*}$\quad Jianglong Ye$^{1}$\quad Jia-Yang Huang$^{1}$\quad Changwei Jing$^{1}$\\ Rocky Duan$^{2}$\quad Pieter Abbeel$^{2,3}$\quad Xiaolong Wang$^{1\dag{}}$\quad Xueyan Zou$^{1\dag{}}$\\
$^1$UC San Diego\quad $^2$Amazon FAR \quad  $^3$UC Berkeley \quad $^*$ Equal Contribution \quad $^{\dag{}}$Equal Advising
\\
\href{https://xl-vla.github.io}{https://xl-vla.github.io}
}

\begin{document}

\twocolumn[{%
\renewcommand\twocolumn[1][]{#1}%
\maketitle
\begin{center}
    \vspace{-20pt}
    \centering
    \captionsetup{type=figure}
    \includegraphics[width=\linewidth]{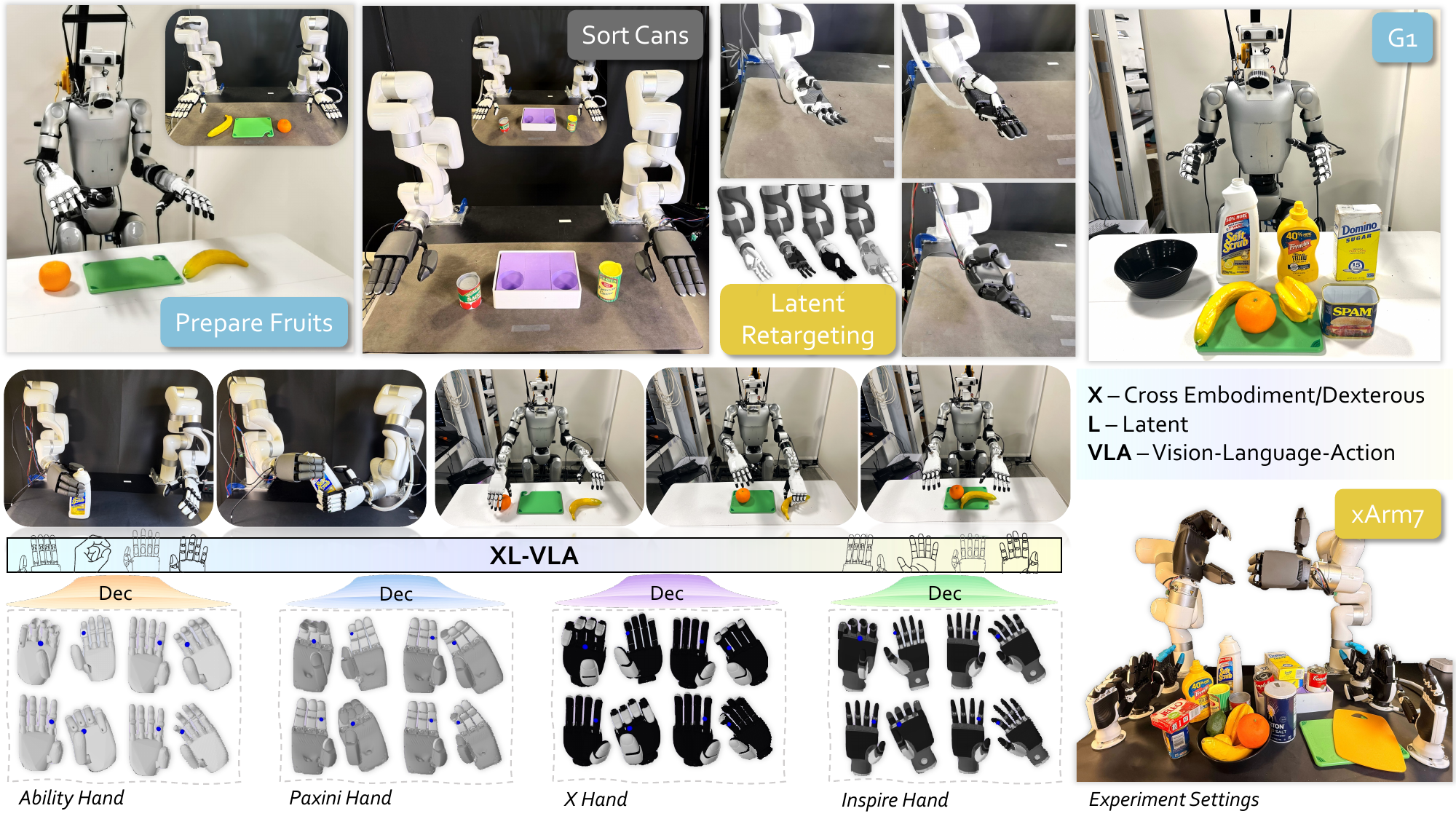}
    \vspace{-1.5em}
    \caption{\textbf{Overview.} \ourmethod{} enables direct decoding of a single latent action into multiple dexterous hand embodiments. Shown above, an action prediction can be instantiated on the Ability hand, Paxini DexH13 hand, X-Hand1, and Inspire hand for language-guided manipulation. We show our experiment settings on the right figure with collected objects and DexHands.}
    \label{fig:teaser}
\end{center}
}]

\maketitle
\begin{abstract}
\vspace{-20pt}

Dexterous manipulation is essential for real-world robot autonomy, mirroring the central role of human hand coordination in daily activity. Humans rely on rich multimodal perception—vision, sound, and language-guided intent—to perform dexterous actions, motivating vision-based, language-conditioned manipulation systems for robots. However, training reliable vision-language-action (VLA) models for dexterous manipulation requires large-scale demonstrations across many robotic hands. In addition, as new dexterous embodiments appear rapidly, collecting data for each becomes costly and impractical, creating a need for scalable cross-embodiment learning. We introduce \ourmethod, a vision-language-action framework integrated with a unified latent action space shared across diverse dexterous hands. This embodiment-invariant latent space is directly pluggable into standard VLA architectures, enabling seamless cross-embodiment training and efficient reuse of both existing and newly collected data. Experimental results demonstrate that \ourmethod consistently outperforms baseline VLA models operating in raw joint spaces, establishing it as an effective solution for scalable cross-embodiment dexterous manipulation.

\end{abstract}    
\vspace{-10pt}
\section{Introduction}
\begin{table*}[t]
\centering
\small
\setlength{\tabcolsep}{3pt}
\renewcommand{\arraystretch}{1.05}

% 1: Paper        2: Data           3: Embodiment      4: EEF<->EEF (elastic)
% 5-9: small centered flag columns
\begin{tabularx}{\textwidth}{
  @{} >{\raggedright\arraybackslash}p{0.13\linewidth}  % Paper
      >{\raggedright\arraybackslash}p{0.23\linewidth}  % Data
      >{\raggedright\arraybackslash}p{0.17\linewidth}  % Embodiment
      >{\raggedright\arraybackslash}X                  % EEF <-> EEF (elastic)
      >{\centering\arraybackslash}p{0.045\linewidth}    % Vision
      >{\centering\arraybackslash}p{0.035\linewidth}    % Lang
      >{\centering\arraybackslash}p{0.035\linewidth}    % Prop
      >{\centering\arraybackslash}p{0.05\linewidth}    % Decoder
      >{\centering\arraybackslash}p{0.05\linewidth}@{}}% ZS
\toprule
\multirow{2}{*}{\textbf{Paper}} &
\multicolumn{1}{c}{\textbf{Data}} &
\multicolumn{2}{c}{\textbf{Deployment}} &
\multicolumn{3}{c}{\textbf{Input}} &
\multicolumn{2}{c}{\textbf{Output}} \\
\cmidrule(lr){2-2}\cmidrule(lr){3-4}\cmidrule(lr){5-7}\cmidrule(lr){8-9}
& \textbf{Type(s)} & \textbf{Embodiment} & \textbf{EEF\,$\leftrightarrow$\,EEF}
& \textbf{Vision} & \textbf{Lang} & \textbf{Prop} & \textbf{Decoder} & \textbf{ZS} \\
\midrule
\textbf{UniVLA}~\cite{BuQ-RSS-25} &
human video; teleoperation &
1 arm+1 gripper &
gripper\,$\leftrightarrow$\,gripper &
\cmark & \cmark & \xmark & \cmark & \xmark \\
\textbf{ATE}~\cite{zhang2025ate} &
teleoperation; simulation &
2 arms+2 grippers &
--- &
\cmark & \cmark & \xmark & \xmark & \xmark \\
\textbf{LAD}~\cite{bauer2025latentdiff} &
teleoperation &
1 arm+1 hand/gripper &
hand\,$\leftrightarrow$\,hand/gripper &
\cmark & \xmark & \xmark & \cmark & \xmark \\
\textbf{EgoBridge}~\cite{punamiya2025egobridge} &
human video; teleoperation &
2 arms+2 pushers &
--- &
\cmark & \xmark & \cmark & \xmark & \xmark \\
\textbf{CoMo}~\cite{Yang2025CoMo} &
internet video; teleoperation &
1 arm+1 gripper &
--- &
\cmark & \xmark & \xmark & \xmark & \xmark \\
\textbf{Tenma}~\cite{davies2025tenma} &
teleoperation &
2 arms+2 grippers &
--- &
\cmark & \cmark & \cmark & \xmark & \xmark \\
\textbf{CycleVAE}~\cite{Dastider2025CycleVAEHBT} &
teleoperation &
1 arm+1 hand &
hand\,$\leftrightarrow$\,hand &
\xmark & \xmark & \cmark & \cmark & \xmark \\
\textbf{CETransfer}~\cite{Wang2024LatentAlignment} &
simulation (sim\,$\to$\,real) &
1 arm+1 gripper &
gripper\,$\leftrightarrow$\,gripper &
\xmark & \xmark & \cmark & \cmark & \cmark \\
\rowcolor{rowhl}
\textbf{Ours} &
teleoperation &
2 arm+2 hand &
hand\,$\leftrightarrow$\,hand &
\cmark & \cmark & \cmark & \cmark & \cmark \\

\bottomrule
\end{tabularx}
\vspace{-5pt}
\caption{\textbf{Related Work Summary.} Summary of related work comparing data sources, deployment settings, and input/output capabilities for latent-based cross-embodiment methods. Data indicates the training modalities used in each work. Deployment specifies the robot embodiments evaluated and whether cross–end-effector transfer is supported. Input denotes which modalities (vision, language, proprioception) are used for training. Output reports whether a method includes a cross-embodiment decoder and whether it enables zero-shot transfer to unseen embodiments.}
\vspace{-12pt}
\label{tab:deployment_alignment}
\end{table*}
Recent progress in vision-language-action (VLA) modeling has begun to extend the successes of large-scale vision and language models into robotics, enabling robots to interpret visual scenes, follow natural language instructions, and execute complex behaviors in the physical world. A key insight behind these advances is that unifying vision and language can be naturally expressed through sequence-to-sequence modeling, and VLA systems can adopt the same abstraction by treating actions as an additional output modality.

However, a fundamental obstacle emerges when moving from vision and language to action: while language possesses a relatively stable and universal vocabulary, robotic action spaces are inherently tied to the morphology of the robot. For dexterous hands in particular, action parameterizations---joint positions---vary significantly across embodiments and continue to evolve rapidly with new hardware designs. This raises two key questions for scalable robot learning: (1) \textit{How can we define a unified action representation within a family of robots?} (2) \textit{How can we seamlessly integrate a new robot whose action space differs from existing ones?}

In this work, we address these challenges by introducing a \emph{shared latent action space} tailored for \textbf{dexterous hands}. This latent space serves as an embodiment-invariant representation that enables joint training across heterogeneous hands. While prior VLA and cross-embodiment efforts have primarily focused on robotic arms equipped with parallel grippers, we focus on the substantially more complex, and more capable domain of dexterous manipulation. Moreover, we emphasize \emph{real-world} datasets and physical robot evaluation, demonstrating that our method remains robust under significant cross-embodiment variation.

\noindent
We summarize our contributions as follows:
\begin{itemize}
    \item We collect a large-scale teleoperation dataset covering 10 manipulation tasks across four newly introduced dexterous hands---Ability, Paxini DexH13, X-Hand1, and Inspire---containing 2M state-action pairs.
    \item We propose an unsupervised latent autoencoder framework that learns a unified action space applicable to a wide range of hands.
    \item We introduce \ourmethod{}, a full VLA pipeline built upon the cross-embodiment latent action space. \ourmethod{} achieves significantly stronger cross-embodiment performance than standard VLA baselines and exhibits \textbf{zero-shot generalization} to untrained cross-embodiment task configurations.
\end{itemize}

\section{Related Work}
\noindent
\textbf{Dexterous Manipulation.} 
The direction of dexterous manipulation focuses on utilizing DexHand for standard manipulation tasks, aiming to enable more complex operations. This field encompasses various areas of focus, including manipulator hardware~\cite{richardson2024isyhand,kuroda2024plexus, ye2025power}, sensors~\cite{sou2024moiretac,xu2024multimodal}, learning and control algorithms~\cite{liu2024dexndm,hung2024avo,chen2024exploiting, jing2026contact}, and human-robot interaction~\cite{chi2024open,fang2024dexop,xu2024dexumi}. In this work, we specifically concentrate on learning and control algorithms, leveraging vision-language-action (VLA) models~\cite{li2024scalable,qu2024eo,yu2024forcevla,song2024ceed, jiang2025gsworld}. Furthermore, we define a unified action space to support cross-embodiment dexterous manipulation~\cite{wei2024omnidexgrasp,fei2024tro,puang2024pchands}.

\begin{figure*}[t]
    \centering
    \includegraphics[width=1.\linewidth]{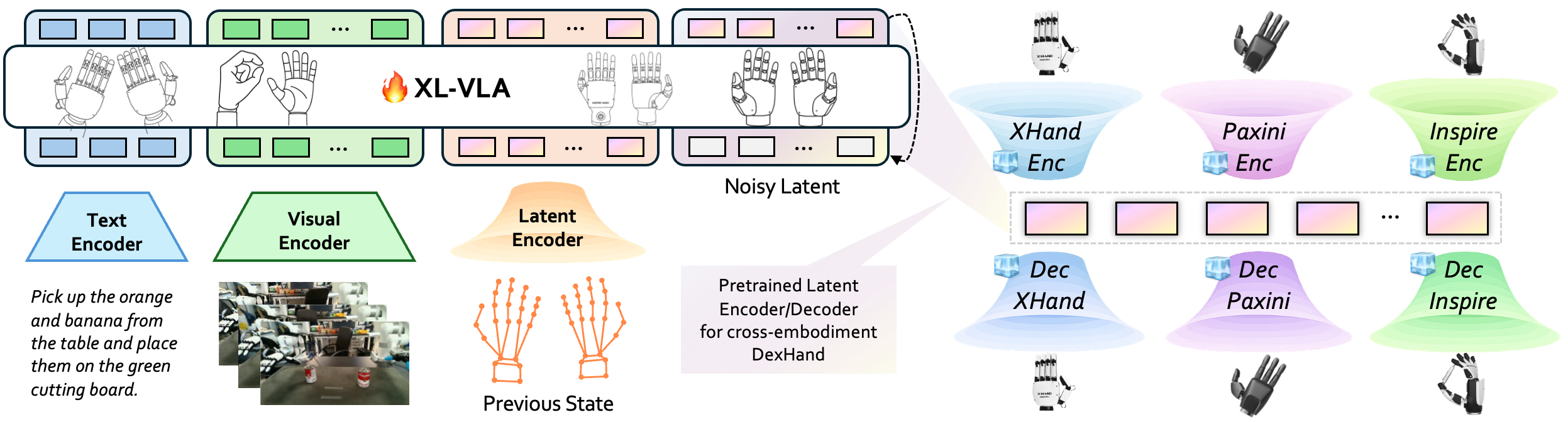}
    \caption{\textbf{Model Pipeline.} \ourmethod{} builds on $\pi_0$~\cite{pi0} with vision and language encoders paired with an action expert that operates in a shared latent action space for cross-embodiment control. During VLA training, the action expert is finetuned while the pretrained latent encoders and decoders remain frozen.}
    \vspace{-10pt}
    \label{fig:method}
\end{figure*}

\noindent
\textbf{Cross Embodiment.}
Cross embodiment typically refers to learning a \emph{single} policy that can flexibly adapt across diverse embodiments—e.g., different humanoids or dexterous hands—without per-robot retraining \cite{tan2025blm1,zheng2025xvla,hou2025dita,doshi2024crossformer,eftekhar2024ring,yang2025multiloco,castro2025vamos, pan2024roboduet}. Within this area, approaches leverage human video for supervision \cite{zakka2021xirl,ren2025motiontracks,dan2025xsim,kim2025uniskill,punamiya2025egobridge,niu2025human2locoman}, apply imitation learning with motion retargeting to bridge morphology gaps \cite{seo2024legato,cao2025gdream,liu2025trajbooster,wu2025cedex}, and employ generative models to synthesize action-consistent trajectories across bodies \cite{bauer2025latentdiff,wen2025dexvla,bai2025roboswap,zhi2025flowaction, huang2024diffusion}. A complementary line of work constructs unified latent action spaces that factor out embodiment-specific details, enabling transfer and zero-shot reuse across platforms \cite{zheng2025uniact,zhang2025ate,bu2025univla,bauer2025latentdiff,wang2024latentalign,davies2025tenma}. This paper follows the latter paradigm, aligning actions in a shared representation for robust cross-embodiment control.

\noindent
\textbf{Hand/Dex Retargeting.} Hand retargeting for teleoperation and imitation learning has progressed from kinematic pipelines to fast, principled learning: GeoRT delivers 1 kHz unsupervised mapping \citep{yin2025geort}; contact-aware and unified formulations improve human–object fidelity \citep{lakshmipathy2024kinematic,lin2025dexflow}, with objective ablations \citep{xin2025analyzing} and practical systems spanning real-time teleop and hardware-agnostic platforms \citep{wen2025bytedexter,chi2025openteledex,dong2025gex}. Beyond copying humans, functional and policy-centric retargeting improves task outcomes \citep{mandi2025dexmachina,xu2025dexplore}, while type-guided teleop exploits robot-specific dexterity \citep{lin2025typetele}.

\noindent
\textbf{Latent Action Space.}
As shown in Table.~\ref{tab:deployment_alignment}, latent action spaces provide embodiment agnostic control codes that align input modalities (vision, language, proprioception) and decode to diverse robots. Examples span discrete VQ tokens with per-robot decoders \cite{BuQ-RSS-25}, continuous end-effector latents trained on retargeted pairs and generated by diffusion \cite{bauer2025latentdiff}, and unified action VAEs that steer existing VLA policies \cite{zhang2025ate}. Other directions align policy features or motion rather than a single EEF space optimal-transport co-training \cite{punamiya2025egobridge}, Internet-video motion embeddings \cite{Yang2025CoMo}, diffusion transformers with standardized tokens \cite{davies2025tenma} and cycle/adversarial mappings enabling cross-embodiment decoding and sim→real transfer \cite{Dastider2025CycleVAEHBT,Wang2024LatentAlignment}.

\noindent
\textbf{Vision-Language-Action Model.} VLA models adapt large vision–language models (VLMs) to robot control by discretizing actions and predicting them autoregressively, enabling transfer of web-scale priors to manipulation \citep{brohan2023rt2visionlanguageactionmodelstransfer, kim2024openvla, wen2024tinyvlafastdataefficientvisionlanguageaction, jiang2024robots}. While these systems demonstrate broad generalization, their tokenized action decoding can hinder high-rate, dexterous control \citep{zhao2023learning}. In contrast, we fine-tune a pre-trained VLM backbone initialized from PaliGemma \citep{beyer2024paligemma} on teleoperated trajectories. Our action expert regresses continuous \emph{latent action chunks}: each target is represented by a single latent vector produced by our hand-specific encoder. During training, we replace $\pi_0$'s original state tokens with these latent tokens and finetune on the next latent chunk, allowing a single hand-agnostic VLA policy to operate across multiple dexterous hands while preserving the benefits of VLM pretraining.

\section{Method}
\subsection{Preliminary}
\noindent
\textbf{Problem Formulation.}
In this work, we address the problem of \textit{language-guided cross-embodiment dexterous manipulation} based on visual perception. For a dexterous hand $h \in \mathcal{H}$ with $d_h$ actuated joints we control absolute joint rotations $\mathbf{q}^{(h)} \in \mathbb{R}^{d_h}$. At the policy level we operate on \emph{action chunks}: each action $\mathbf{q}_t^{(h)} \in \mathbb{R}^{64 \times d_h}$ is a sequence of $64$ joint-position commands sampled at $20$\,Hz (3.2\,s of motion). At control step $t$, the policy receives a short history of joint states, the previously executed action chunk $\mathbf{q}_t^{(h)}$, the current image observations $\mathbf{V}$, and a language instruction $\mathbf{T}$, and predicts the next chunk $\mathbf{q}_{t+1}^{(h)}$ via
$\mathbf{q}_{t+1}^{(h)} = F(\mathbf{q}_t^{(h)}, \mathbf{V}, \mathbf{T})$.

The objective is to predict embodiment-consistent joint-rotation trajectories conditioned on these multimodal inputs with a unified multi-task VLA model. Although the continuous joint spaces $\mathbf{q}^{(h)}$ are hand-specific, the sequence model $F$ itself is hand-agnostic; the hand identity $h$ is used only to choose the appropriate encoder/decoder that maps between joint space and the shared latent action space described below. To evaluate this setting, we consider a diverse set of dexterous robotic hands, including the \textit{Ability Hand}, \textit{Inspire Hand}, \textit{X-Hand1}, and \textit{Paxini DexH13}, which vary in structure, actuation, and kinematics.

To tackle this problem, we introduce an embodiment-invariant latent action space that integrates seamlessly into a vision–language–action (VLA) framework. This latent space provides a unified representation across diverse dexterous hands, enabling the model to train effectively on cross-embodiment data and generalize manipulation skills beyond a single hand morphology. Furthermore, the proposed latent space supports transferring control policies across different embodiments without requiring hand-specific retraining.

\noindent
\textbf{Pipeline.}
As illustrated in Fig.~\ref{fig:method}, our proposed framework consists of two main components:
(1) a VLA backbone that encodes multimodal inputs $(\mathbf{V}, \mathbf{T})$, and (2) a pretrained set of latent encoders and decoders designed for cross-embodiment transfer. Our VLA design follows $\pi_0$~\cite{pi0}, which employs vision and language encoders together with an action expert. In the original $\pi_0$, proprioceptive history is provided through a stack of state tokens. In XL-VLA we instead feed \emph{latent action tokens}: for each hand $h$, a hand-specific encoder $E_h$ maps the previous absolute joint-position action chunk $\mathbf{q}_t^{(h)}$ (64 frames at $20$\,Hz) into a compact latent vector $\mathbf{z}_t = E_{h}(\mathbf{q}_t^{(h)})$. The VLA model conditions on a short history of such latent tokens, together with vision and language tokens, and predicts the next latent chunk $\widehat{\mathbf{z}}_{t+1}$. This latent is decoded by the embodiment-specific decoder $D_{h}$ to obtain the next joint command chunk $\widehat{\mathbf{q}}_{t+1}^{(h)} = D_{h}(\widehat{\mathbf{z}}_{t+1})$. During VLA finetuning we keep all latent encoders and decoders frozen.

This embodiment-invariant latent representation $\mathbf{z}$ acts as a unified action space shared across heterogeneous dexterous hands. By learning and decoding actions within this latent space, the model effectively bridges differences in morphology and actuation across embodiments, enabling a single hand-agnostic VLA policy to operate on diverse robotic hands. The hand identity $h$ is used only to select the appropriate encoder $E_h$ and decoder $D_h$ and is never provided as an explicit input token to the VLA backbone. We describe the detailed design and training of this latent action space in the following sections.

\subsection{Latent Space}

\noindent
\textbf{Definition.}
Rather than defining a separate action space for each dexterous hand, we introduce a \emph{shared latent action space} that provides a unified representation for all dexhand embodiments. This latent space is pretrained independently of the VLA model through a set of hand-specific encoders and decoders that all map to the same latent distribution. As a result, the latent embedding acts as an implicit, embodiment-agnostic action space that can be used by downstream policies to seamlessly control different dexterous hands.

\subsubsection{Modeling}

To construct the latent representation, we employ a multi-headed VAE-style autoencoder. 
For each hand type $h \in \mathcal{H}$ (e.g., X-Hand, Ability, Inspire, Paxini), 
we define a hand-specific encoder $E_h$ and decoder $D_h$. 
Each hand provides a joint configuration $\mathbf{q}^{(h)} \in \mathbb{R}^{d_h}$,
where $\mathbf{q}^{(h)}$ denotes the joint position values (q-pos) and $d_h$ is the dimensionality of that embodiment. The encoder outputs the parameters of a Gaussian posterior $(\boldsymbol{\mu}^{(h)}, \boldsymbol{\sigma}^{(h)}) = E_h(\mathbf{q}^{(h)})$, from which we sample a latent code $\mathbf{z}$ using the reparameterization trick, $q(\mathbf{z}\mid\mathbf{q}^{(h)}) = \mathcal{N}(\boldsymbol{\mu}^{(h)}, \mathrm{diag}((\boldsymbol{\sigma}^{(h)})^2))$. The decoder reconstructs back into the corresponding joint space $\hat{\mathbf{q}}^{(h)} = D_h(\mathbf{z})$.

In practice, each encoder and decoder is implemented as a lightweight MLP:  
the input q-pos vector $\mathbf{q}^{(h)}$ is projected into a common latent space through the encoder MLP, and the decoder MLP reprojects the latent embedding back into the hand's original joint configuration. This architecture provides a unified latent manifold while preserving the structure of each embodiment. To shape a meaningful cross-embodiment latent space, we impose three training constraints: (1) a \emph{reconstruction constraint} $L_1$ ensuring $\hat{\mathbf{q}}^{(h)}$ matches $\mathbf{q}^{(h)}$, (2) a \emph{retargeting constraint} $L_2$ aligning fingertip geometry across hands using differentiable forward kinematics,  
and (3) a \emph{latent constraint} $L_3$ regularizing the latent embedding to follow a smooth prior distribution. Together, these constraints encourage the latent space to capture embodiment-invariant structure, enabling consistent decoding into any dexterous hand.

\subsubsection{Objective}
\noindent\textbf{Reconstruction Loss ($L_1$).}
Since each DexHand embodiment has its own joint space, we first require that the
hand-specific encoder-decoder pair behaves as an autoencoder on that hand.  
Given a joint configuration $\mathbf{q}^{(h)}$ and its reconstruction 
$\hat{\mathbf{q}}^{(h)}$ for hand $h \in \mathcal{H}$, the reconstruction loss
averaged over all hands is
\begin{equation}
    L_1 = \mathcal{L}_{\mathrm{rec}}
    =
    \frac{1}{|\mathcal{H}|}
    \sum_{h \in \mathcal{H}}
    \mathrm{MSE}\!\left(\hat{\mathbf{q}}^{(h)}, \mathbf{q}^{(h)}\right),
\end{equation}
which ensures that the latent space preserves hand-specific kinematics and that
no embodiment is degraded by sharing the latent representation.

\begin{table*}[ht]
\centering
\small
\setlength{\tabcolsep}{3pt}
\renewcommand{\arraystretch}{1.1}

\begin{tabularx}{\textwidth}{
  @{}>{\raggedright\arraybackslash}p{0.08\linewidth}
  >{\raggedright\arraybackslash}p{0.06\linewidth}
  *{11}{>{\centering\arraybackslash}X}@{}
}
\toprule

% ===== Header Row =====
\textbf{Method} & \textbf{Hand} &
\textbf{PF} & \textbf{SC} & \textbf{SoC} & \textbf{HB} & \textbf{RL} &
\textbf{PS} & \textbf{RB} & \textbf{PuS} & \textbf{PoS} & \textbf{PC} &
\textbf{Mean} \\

\midrule

% ===== Body: 2 main groups × 4 subrows each =====
\multirow{4}{*}{$\pi_0$~\cite{pi0}} &
Ability &
0.10 &
0.10 &
0.00 &
0.70 &
0.20 &
0.80 &
0.60 &
0.30 &
0.30 &
0.60 &
0.37 \\

& Inspire &
0.10 &
0.20 &
0.00 &
0.30 &
0.10 &
0.50 &
0.30 &
0.20 &
0.20 &
0.80 &
0.27 \\

& Paxini &
0.40 &
0.40 &
0.30 &
0.20 &
0.00 &
0.80 &
0.60 &
0.30 &
0.10 &
0.40 &
0.35 \\

& XHand &
0.20 &
0.40 &
0.00 &
0.40 &
0.10 &
0.60 &
0.30 &
0.20 &
0.30 &
0.40 &
0.29 \\

\rowcolor{rowhl}
& Mean &
0.20 & 0.28 & 0.08 & 0.40 & 0.10 & 0.68 & 0.45 & 0.25 & 0.23 & 0.55 & 0.32 \\

\midrule

\multirow{4}{*}{\ourmethod} &
Ability &
\textbf{0.80} &
\textbf{0.80} &
0.40 &
\textbf{1.00} &
0.70 &
\textbf{1.00} &
0.70 &
0.30 &
0.90 &
0.70 &
0.73 \\

& Inspire &
0.60 &
0.50 &
0.50 &
0.80 &
0.40 &
0.80 &
\textbf{1.00} &
\textbf{0.40} &
0.80 &
\textbf{1.00} &
0.68 \\

& Paxini &
\textbf{0.80} &
0.70 &
\textbf{0.80} &
\textbf{1.00} &
0.30 &
\textbf{1.00} &
\textbf{1.00} &
\textbf{0.40} &
0.80 &
\textbf{1.00} &
\textbf{0.78} \\

& XHand &
0.60 &
0.50 &
0.50 &
\textbf{1.00} &
0.30 &
\textbf{1.00} &
0.90 &
0.30 &
\textbf{1.00} &
0.90 &
0.70 \\

\rowcolor{rowhl}
& Mean &
0.70\impr{50\%} & 0.63\impr{35\%} & 0.55\impr{47\%} & 0.95\impr{55\%} & 0.43\impr{33\%} &
0.95\impr{27\%} & 0.90\impr{45\%} & 0.35\impr{10\%} & 0.88\impr{65\%} & 0.90\impr{35\%} & 0.72\impr{40\%} \\

\bottomrule
\end{tabularx}
\vspace{-3pt}
\caption{\textbf{Vision-Language-Action Modeling.} We compare \ourmethod{} with $\pi_0$ under cross-embodiment training. Although $\pi_0$ can handle different embodiments by adjusting sequence length, our method achieves consistently higher success rates across all hands and tasks. The first row PF, SC, SoC, etc. denote each task introduced in Task \& Dataset, and each task is executed for 10 times, and we compute the success rate for each task cross hand.}
\vspace{-10pt}
\label{tab:main}
\end{table*}

% \begin{table*}[ht]
% \centering
% \small
% \setlength{\tabcolsep}{3pt}
% \renewcommand{\arraystretch}{1.1}

% \begin{tabularx}{\textwidth}{
%   @{}>{\raggedright\arraybackslash}p{0.08\linewidth}
%   >{\raggedright\arraybackslash}p{0.06\linewidth}
%   *{11}{>{\centering\arraybackslash}X}@{}
% }
% \toprule

% % ===== Header Row =====
% \textbf{Method} & \textbf{Hand} &
% \textbf{Task 1} & \textbf{Task 2} & \textbf{Task 3} &
% \textbf{Task 4} & \textbf{Task 5} & \textbf{Task 6} &
% \textbf{Task 7} & \textbf{Task 8} & \textbf{Task 9} &
% \textbf{Task 10} \\

% \midrule

% % ===== Body: 2 main groups × 4 subrows each =====
% \multirow{4}{*}{$\pi_0$~\cite{pi0}} &
% Ability & 10 & 10 & 0 & 70 & 20 & 80 & 60 & 30 & 30 & 60 \\
% & Inspire & 10 & 20 & 0 & 30 & 10 & 50 & 30 & 20 & 20 & 80 \\
% & Paxini & 40 & 40 & 30 & 20 & 0 & 80 & 60 & 30 & 10 & 40 \\
% & XHand  & 20 & 40 & 0 & 40 & 10 & 60 & 30 & 20 & 30 & 40 \\

% \midrule

% \multirow{4}{*}{\ourmethod} &
% Ability & 80 & 80 & 40 & 100 & 70 & 100 & 70 & 30 & 90 & 70 \\
% & Inspire & 60 & 50 & 50 & 80 & 40 & 80 & 100 & 40 & 80 & 100 \\
% & Paxini & 80 & 70 & 80 & 100 & 30 & 100 & 100 & 40 & 80 & 100 \\
% & XHand  & 60 & 50 & 50 & 100 & 30 & 100 & 90 & 30 & 100 & 90 \\

% \bottomrule
% \end{tabularx}

% \caption{Main.}
% \label{tab:main}
% \end{table*}

\begin{figure}[t]
    \centering
    \includegraphics[width=1.\linewidth]{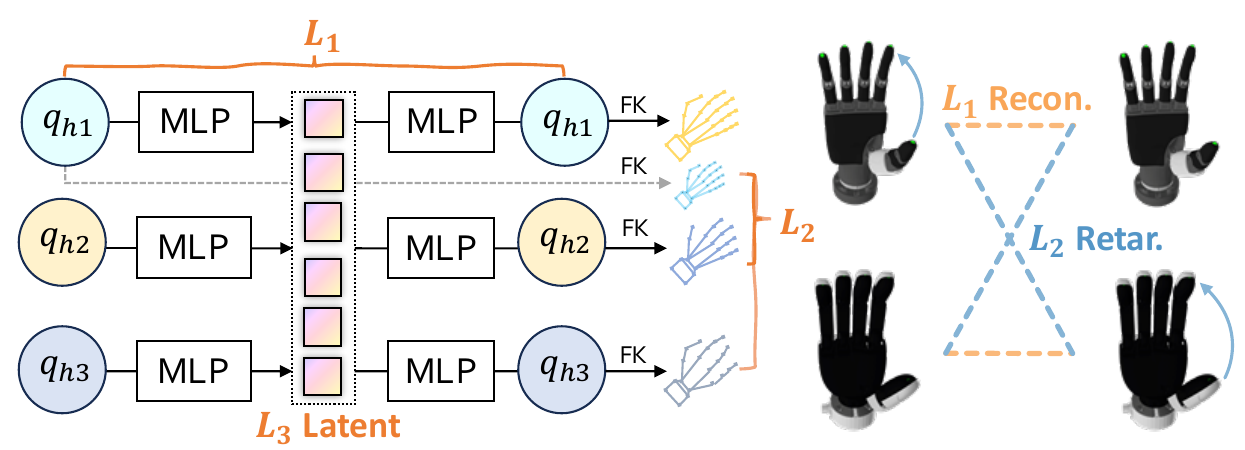}
    \vspace{-20pt}
    \caption{\textbf{Latent space pretraining pipeline.} For each hand type, joint positions $\mathbf{q}_{h}$ are mapped through an encoder MLP into a shared latent space and reconstructed by a decoder MLP. The diagram also indicates the placement of the reconstruction loss $L_1$, retargeting loss $L_2$ via differentiable forward kinematics, and latent regularization loss $L_3$.}
    \vspace{-15pt}
    \label{fig:latent_ill}
\end{figure}

\vspace{4pt}
\noindent\textbf{Retargeting Loss ($L_2$).}
To make the latent space truly cross-embodiment, we align fingertip geometry
between different DexHand robots.  
For each hand $h$, we use differentiable forward kinematics (FK) to map joints
to fingertip positions $\mathbf{p}^{(h)}_i$, and define fingertip
displacements $\boldsymbol{\delta}^{(h)}_{ij} = \mathbf{p}^{(h)}_i - \mathbf{p}^{(h)}_j$ for fingertip pairs $(i,j) \in \mathcal{P}$. The pair set $\mathcal{P}$ contains thumb--finger pairs for the four aligned digits (thumb--index, thumb--middle, thumb--ring, thumb--little). We manually align finger indices across hands so that these digits correspond semantically; for Paxini DexH13, which lacks a little finger, we drop any pairs involving that digit when evaluating $L_2$. The retargeting loss penalizes discrepancies in pinch distances and directions between source hands $s$ and target hands $t$:
\vspace{-3pt}
{\scriptsize
  \begin{equation}
  \begin{aligned}
  L_2
  &= \frac{1}{|\mathcal{H}|(|\mathcal{H}|-1)|\mathcal{P}|}
     \sum_{s \neq t}
     \sum_{(i,j)\in\mathcal{P}}
  % &\quad
     w_{ij}^{(s)}
     \bigg[
        \lambda_{\mathrm{dis}}
        \big(
           \|\boldsymbol{\delta}_{ij}^{(s)}\|_2 -
           \|\hat{\boldsymbol{\delta}}_{ij}^{(t)}\|_2
        \big)^2 \\
  &\qquad\qquad
        +
        \lambda_{\mathrm{dir}}
        \big(
           1 - c_{ij}^{(s,t)}
        \big)
     \bigg].
  \end{aligned}
  \end{equation}
}
where $\hat{\boldsymbol{\delta}}^{(t)}_{ij}$ is computed from the decoded
configuration of hand $t$, $c_{ij}^{(s,t)}$ denotes the cosine of the angle between the pinch directions $\boldsymbol{\delta}_{ij}^{(s)}$ and $\hat{\boldsymbol{\delta}}_{ij}^{(t)}$, and
$w_{ij}^{(s)} = \exp(-\lambda_{\mathrm{dis}}^{\mathrm{exp}}
\|\boldsymbol{\delta}_{ij}^{(s)}\|_2)$ emphasizes tighter pinches.  
This loss encourages the same latent code to produce geometrically consistent
pinch behaviors across different hands.

\medskip
\noindent\textbf{Latent Loss ($L_3$).}
Finally, we regularize the DexHand latent space to be smooth and well-behaved
by imposing a standard Gaussian prior on the latent variables.  
For the approximate posterior $q(\mathbf{z}\mid\mathbf{q})$ produced by the
hand-specific encoders, the latent loss is
\begin{equation}
    L_3 = \mathcal{L}_{\mathrm{KL}}
    =
    \mathbb{E}_{\mathbf{q}}
    \big[
        \mathrm{KL}\big(
            q(\mathbf{z}\mid\mathbf{q}) 
            \,\|\, 
            \mathcal{N}(\mathbf{0}, \mathbf{I})
        \big)
    \big],
\end{equation}
which encourages the shared DexHand latent space to follow a 
$\mathcal{N}(0, I)$ distribution and facilitates sampling and interpolation
across embodiments.

\medskip
\noindent\textbf{Training Data and Protocol.}
The latent autoencoder is trained \emph{without} any demonstration or IK-generated trajectories. Instead, for each hand $s \in \mathcal{H}$ we randomly sample joint configurations within the hardware joint limits to form synthetic joint-position vectors $\mathbf{q}^{(s)}$. For every such sample we encode $\mathbf{q}^{(s)}$ to a latent $\mathbf{z}$, decode it through all decoders $\{D_t\}_{t \in \mathcal{H}}$, and accumulate reconstruction and retargeting losses: the self-decoding $D_s(\mathbf{z})$ contributes to $L_1$, while cross-hand decodings $D_t(\mathbf{z})$ for $t \neq s$ contribute to $L_2$. Losses from all hands are aggregated and a single backward pass is applied, so all encoders and decoders are optimized jointly. Because $L_2$ uses only forward kinematics of each hand and decoded poses, the alignment of the latent space across embodiments is completely self-supervised and does not require any paired cross-hand trajectories.

\medskip
\noindent\textbf{Total Latent Objective.}
The full latent training loss combines reconstruction, retargeting, and KL regularization:
\begin{equation}
    L_{\mathrm{latent}} = L_1 + L_2 + \beta L_3.
\end{equation}
In all experiments we fix the weights to
$\beta = 10^{-5}$,
$\lambda_{\mathrm{dis}} = 2000.0$,
$\lambda_{\mathrm{dir}} = 5.0$, and
$\lambda_{\mathrm{dis}}^{\mathrm{exp}} = 12.0$.
These values yield a latent space that is both geometrically well-aligned across hands and smooth enough to support sampling and interpolation.

\section{Experiments}

\begin{figure*}[t]
    \centering
    \includegraphics[width=\linewidth]{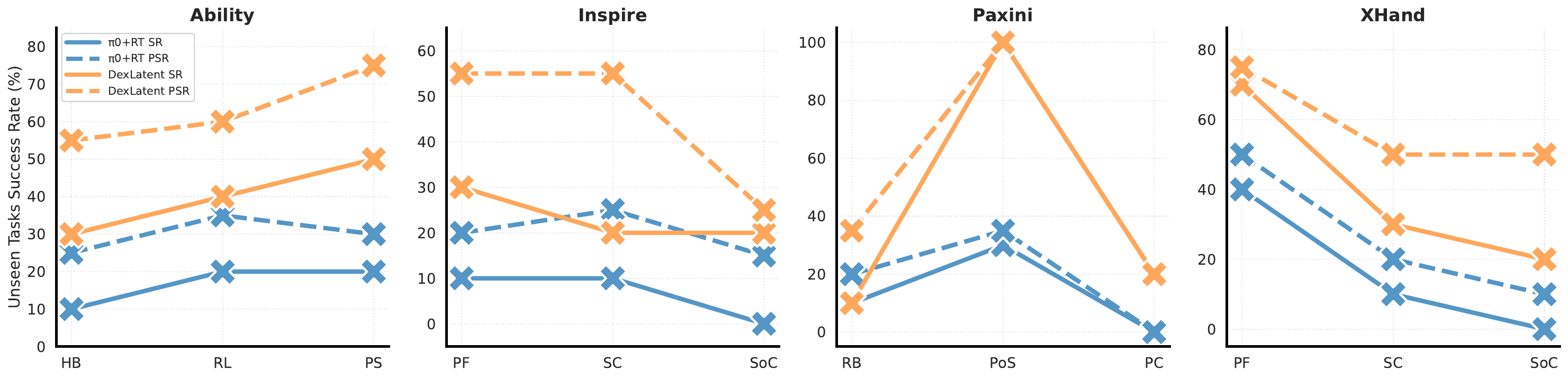}
    \vspace{-18pt}
    \caption{\textbf{Zero-shot Unseen Tasks Generalization}. For each hand, we randomly select some tasks as unseen tasks, whose data are held out from the training dataset. Then we test the unseen tasks with model trained on other data. Results show that by training with an aligned latent action space, \ourmethod{} gets the ability to generalize to novel hand-task combination in a zero-shot manner. PSR stands for ``Partial Success Rate", where policy is rewarded with half success if only one arm finishes its task.}
    \vspace{-12pt}
    \label{fig:unseen_task}
\end{figure*}

\noindent
\textbf{Tasks \& Dataset.}
We design 10 diverse tasks with different skills and objects to evaluate our VLA models. For each task, we collect 50 demonstrations each task per hand via~\cite{bunny-visionpro}, with 2000 demonstrations collected in total. Task descriptions are listed below:

{\small
\noindent (a) Prepare Fruits (PF). Put the banana and orange on the green board for cutting. \\
\noindent (b) Stack Cans (SC). Stack the cheese can on top of the salt. \\
\noindent (c) Sort Cans (SoC). Put the tomato can and the cheese can into the container. \\
\noindent (d) Hand over Bottle (HB). Hand over the white bottle from right hand to left hand. \\
\noindent (e) Re-organize Lemons (RL). Put the yellow lemon and the green lime into the bowl. \\
\noindent (f) Pour Sauce (PS). Pour mustard sauce into the meat can. \\
\noindent (g) Re-arrange Boxes (RB). Keep the table organized by re-arranging the two boxes. \\
\noindent (h) Push Sugar (PuS). Push the sugar boxes together. \\
\noindent (i) Pour Sugar (PoS). Add sugar to the starfruit. \\
\noindent (j) Push Cans (PC). Push the two tomato cans together. \\
\noindent
}

\vspace{-10pt}
\noindent\textbf{Hardware.}
To evaluate our method, we conduct comprehensive experiments on our real-world robot platform. We use a bimanual 7-DoF xArm and a Unitree G1 humanoid with various robot hands, shown in table~\ref{tab:hand-cmp}.

\begin{table}[h]
\vspace{-6pt}
\centering
\footnotesize
\setlength{\tabcolsep}{2pt}  % reduce horizontal padding
\renewcommand{\arraystretch}{0.9} % reduce vertical padding
\begin{tabular}{lccccc}
\toprule
 &
Ability &
Inspire &
X-Hand1 &
Paxini DexH13 \\
\midrule
\#Fingers  & 5 & 5  & 5      & 4         \\
\#DoF(mimic) & 12(6) & 12(6)  & 12    & 16(3)     \\
\bottomrule
\end{tabular}
\vspace{-8pt}
\caption{\textbf{Dexterous Hand Comparison.}}
\label{tab:hand-cmp}
\vspace{-12pt}
\end{table}

\noindent
\textbf{Experiment Settings.} We initialize the \ourmethod with weights from~\cite{pi0}, then train the model on our collected multi-embodiment dataset. We use 8 NVIDIA H100 GPUs to train \ourmethod, each having 80GB memory. The model is trained 60K steps with a batch size of 128. Note that \ourmethod is a unified cross-embodiment multi-task policy. We use language to condition the policy on multiple tasks.

\subsection{Main Results}
In this section, we are trying to answer the following questions that mainly focus on \textit{Effectiveness of VLA + Latent Integration:}
(1) Does \ourmethod{} outperform standard VLA models in cross-embodiment training? (2) Does \ourmethod{} enable zero-shot cross-embodiment skill transfer?

\noindent
\textbf{Cross-Hand Data Scaling.}
Tab.\ref{tab:main} presents the cross-embodiment manipulation results for \ourmethod{} compared with a strong $\pi_0$ baseline trained on the same multi-hand, multi-task dataset as a single shared policy across all four hands—Ability, Inspire, Paxini, and X-Hand—and ten manipulation tasks. Although $\pi_0$ can nominally accommodate different embodiments by varying sequence lengths, its performance remains inconsistent and generally low due to substantial kinematic and actuation differences across hands. In contrast, \ourmethod{} achieves strong and consistent improvements for every hand and task, demonstrating the benefit of learning a shared latent action representation.

Across hands, \ourmethod{} yields notable per-embodiment gains. The Ability Hand, which features relatively simple actuation, benefits from a large boost in reliability, improving from 0.37 to 0.73 overall. The Paxini Hand achieves highest performance among all embodiments (0.78 overall) , indicating strong compatibility between its actuation structure and the learned latent mapping. XHand , which is the most mechanically distinct from the rest, also improves significantly from 0.29 to 0.70, showing that \ourmethod{} can bridge large embodiment gaps.

Averaged over all tasks and hands, \ourmethod{} increases the mean success rate from 0.55 to 0.90 (+0.35). Particularly large improvements are observed for dexterity-heavy tasks such as Sort Cans, Hand over Bottle, and Re-arrange Boxes, underscoring the effectiveness of our embodiment-invariant latent space in capturing fine-grained manipulation behavior. Overall, these results demonstrate that \ourmethod{} enables robust cross-embodiment action prediction and consistently surpasses VLA models that lack a unified action representation.

\begin{figure}[h]
    \centering
    \includegraphics[width=\linewidth]{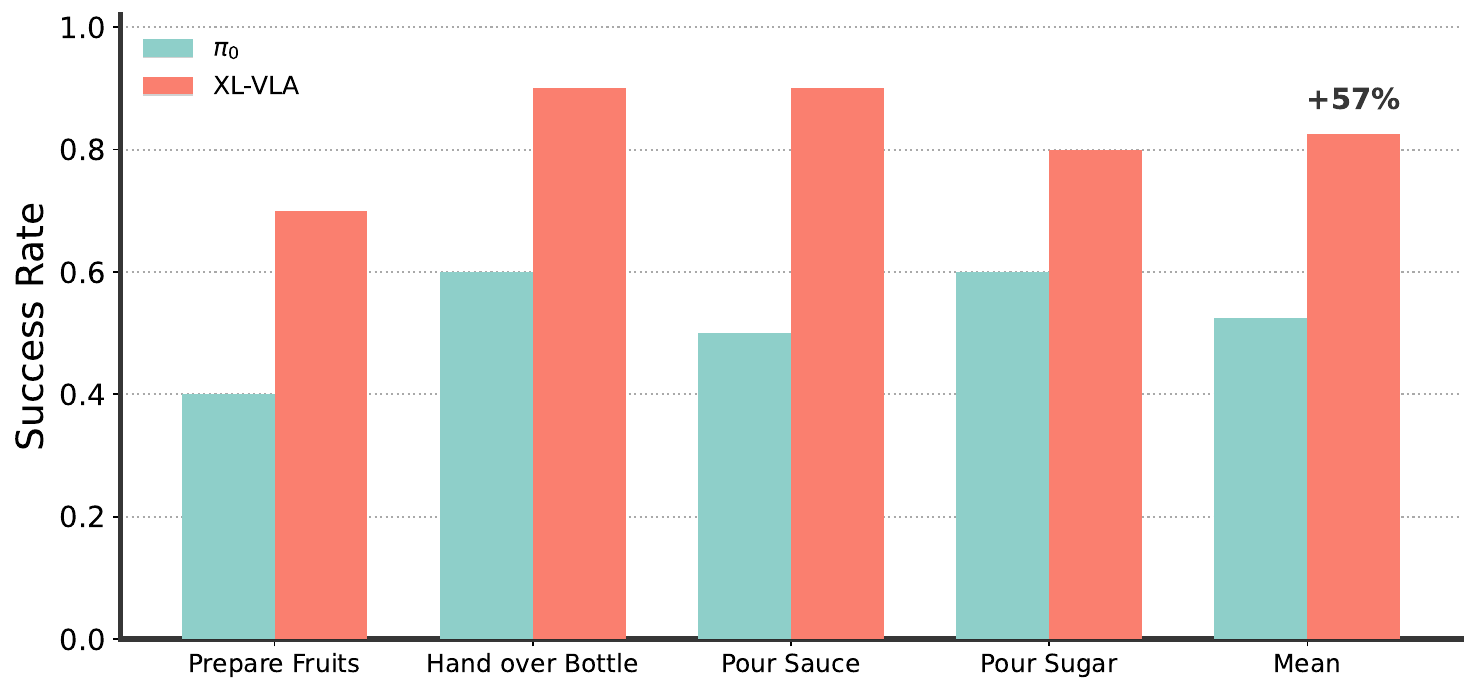}
    \caption{\textbf{G1 Cross-Robot Performance.} Co-training with latent xArm and humanoid data outperforms using raw actions.}
    \label{fig:g1-bar}
    \vspace{-12pt}
\end{figure}

\noindent
\textbf{Cross-Robot Data Scaling.} To show the unified latent space benefit even for different robot systems, we test four manipulation tasks with data from the tabletop xArm and humanoid G1. We co-train the data from all embodiments on the four tasks with the same training parameters and show the G1 success rates in figure~\ref{fig:g1-bar}. We can see that simply using aligned latent action space boost the performance of training on the raw action space, which has varied state/action lengths.

\noindent
\textbf{Zero-Shot Task Generalization.}
A key advantage of using an embodiment-invariant latent action space is its ability to support seamless \textit{zero-shot generalization} to unseen tasks. Because all dexterous hands share the same latent representation, a policy trained on a subset of tasks with one embodiment can transfer to a different task--hand combination without requiring additional training or retargeting. 

\begin{table*}[t]
\centering
\small
\setlength{\tabcolsep}{3pt}
\renewcommand{\arraystretch}{1.1}

\begin{tabularx}{\textwidth}{
  @{}>{\raggedright\arraybackslash}p{0.08\linewidth}
  >{\raggedright\arraybackslash}p{0.12\linewidth}
  *{11}{>{\centering\arraybackslash}X}@{}
}
\toprule

% ===== Header Row =====
\textbf{Model} & \textbf{Combination} &
\textbf{PF} & \textbf{SC} & \textbf{SoC} & \textbf{HB} & \textbf{RL} &
\textbf{PS} & \textbf{RB} & \textbf{PuS} & \textbf{PoS} & \textbf{PC} &
\textbf{Mean} \\

\midrule

% ===== Body: 2 main groups × 4 subrows each =====
\multirow{2}{*}{\textbf{LAD~\cite{bauer2025latentdiff}}} &
Ability+Inspire & 0.8 & 0.4 & 0.5 & 0.4 & 0.6 & 0.6 & 0.6 & 0.5 & 0.7 & 0.9 & 0.60 \\
& Paxini+XHand & 0.7 & 0.5 & 0.6 & 0.7 & 0.5 & 0.8 & 0.7 & 0.4 & 0.6 & 0.6 & 0.61 \\

\midrule

\multirow{2}{*}{\textbf{\ourmethod}} &
Ability+Inspire & \textbf{0.9} & \textbf{0.7} & 0.8 & 0.7 & \textbf{0.8} & 0.7 & 0.8 & \textbf{0.8} & \textbf{1.0} & \textbf{1.0} & \textbf{0.82} \\
& Paxini+XHand & 0.8 & \textbf{0.7} & \textbf{0.9} & \textbf{0.8} & 0.6 & \textbf{0.9} & \textbf{0.9} & 0.6 & \textbf{1.0} & 0.9 & 0.81 \\

\bottomrule

\end{tabularx}
\vspace{-5pt}
\caption{\textbf{Latent replay comparison}. We compare our latent space with Latent Action Diffusion (LAD)~\cite{bauer2025latentdiff}. For each hand combination, teleoperated trajectories collected on one source hand are encoded into the latent space, decoded onto the target hand, and replayed on real hardware. A replay is counted as successful if the encoded--decoded sequence can be executed without breaking contact or causing self-collisions. Higher replay success indicates better cross-embodiment consistency of the latent representation.}
\label{tab:replay_latent_cmp}
\vspace{-10pt}
\end{table*}

To evaluate this capability, we hold out several manipulation tasks as \textit{unseen tasks} for each hand and train \ourmethod{} on the remaining tasks. At test time, the trained policy is applied directly to the unseen task through the corresponding embodiment-specific decoder. As shown in Fig.~\ref{fig:unseen_task}, we report both \textit{absolute success rate} (SR) and \textit{partial success rate} (PSR), where PSR accounts for intermediate progress (0.25, 0.5, 0.75, 1.0) to provide a more fine-grained measure of policy performance.

For comparison, we construct a $\pi_0$+RT baseline in which a policy is trained on all tasks using only the XHand embodiment. During evaluation, we apply a standard \textit{kinematic retargeting} algorithm to map the predicted XHand joint trajectories to the other embodiments (Inspire, Ability, Paxini) by aligning fingertip positions. This baseline reflects common practice in cross-embodiment manipulation and allows us to assess whether our latent action representation provides genuine zero-shot benefits over retargeting-based transfer.

Across all embodiments and tasks, XL-VLA consistently outperforms the retargeting VLA baseline, often by a substantial margin. Notably, XL-VLA \textit{never} underperforms the baseline on any hand or task, highlighting the robustness of the latent action representation. The gains are especially pronounced on fine-grained dexterous tasks (e.g., \texttt{HB}, \texttt{RB}), where geometric retargeting struggles to maintain coordinated finger motion.

\begin{table*}[htbp]
      \centering
      \setlength{\tabcolsep}{3pt}
      \renewcommand{\arraystretch}{1.15}
      \begin{tabularx}{\textwidth}{c *{10}{S[table-format=3.3]}}
          \toprule
          & \multicolumn{2}{c}{\textbf{Reconstruction $\downarrow$}} & 
            \multicolumn{4}{c}{\textbf{Cross Embodiment $\downarrow$}} & 
            \multicolumn{2}{c}{\textbf{Latent Continuity $\downarrow$}} & 
            \multicolumn{2}{c}{\textbf{Interpolation $\downarrow$}} \\
          \cmidrule(lr){2-3}\cmidrule(lr){4-7}\cmidrule(lr){8-9}\cmidrule(lr){10-11}
          \textbf{Exp} & 
          \multicolumn{1}{c}{\textbf{\hspace{4pt}Joint}} & 
          \multicolumn{1}{c}{\textbf{Tip}} & 
          \multicolumn{1}{c}{\textbf{PT$^{\text{dir}}$}} & 
          \multicolumn{1}{c}{\textbf{\hspace{12pt}PT$^{\text{dist}}$}} & 
          \multicolumn{1}{c}{\textbf{RT$^{\text{dir}}$}} & 
          \multicolumn{1}{c}{\textbf{\hspace{8pt}RT$^{\text{dist}}$}} & 
          \multicolumn{1}{c}{\textbf{Joint}} & 
          \multicolumn{1}{c}{\textbf{Tip}} & 
          \multicolumn{1}{c}{\textbf{Accel.}} & 
          \multicolumn{1}{c}{\textbf{Jerk}} \\
          \midrule
          \rowcolor{orange!15}
          Ours & 5.476 & 3.703 & 11.857 & 1.872 & 10.492 & 6.295 & 4.492 & 8.534 & 8.683 & 9.659 \\
          $-L_1$ & {\textcolor{badresult}{\hspace{4pt}61.672}} & 39.400 & 11.741 & 1.857 & 10.398 & 6.375 & 24.784 & 58.858 & 12.028 & 16.852 \\
          $-L_2^{\text{dist}}$ & 5.195 & 3.580 & 3.972 & 4.413 & 6.788 & {\textcolor{badresult}{\hspace{6pt}24.488}} & 4.073 & 8.168 & 8.525 & 9.240 \\
          $-L_2^{\text{dir}}$ & 4.966 & 3.378 & 46.217 & 2.251 & {\textcolor{badresult}{\hspace{4pt}53.546}} & 5.518 & 4.451 & 9.217 & 8.742 & 9.551 \\
          $-L_2$ (both) & 3.781 & 2.602 & 62.733 & 8.080 & {\textcolor{badresult}{\hspace{4pt}71.765}} & 62.809 & 2.823 & 6.757 & 8.602 & 9.426 \\
          $H^{128}_{256}$ & 5.897 & 3.908 & 9.073 & 1.613 & 10.432 & 6.277 & 3.104 & 6.410 & 9.213 & {\textcolor{badresult}{\hspace{6pt}10.406}} \\
          $H^{64}_{128} \times 2$ & {\textcolor{badresult}{\hspace{10pt}8.216}} & 4.280 & 9.027 & 1.513 & 10.572 & 6.713 & 5.004 & 8.832 & 8.559 & 9.479 \\
          $H^{64}_{64}$ & 4.979 & 3.411 & 9.010 & 1.655 & 10.702 & {\textcolor{badresult}{\hspace{10pt}6.985}} & 2.922 & 6.298 & 8.618 & 9.296 \\
          $H^{64}$ & 5.021 & 3.445 & 9.010 & 1.518 & 10.213 & {\textcolor{badresult}{\hspace{10pt}6.435}} & 4.174 & 8.132 & 8.246 & 8.740 \\
          $\mathbf{L}_{8}$ & {\textcolor{badresult}{\hspace{4pt}20.913}} & 6.499 & 9.217 & 1.557 & 10.960 & 6.805 & 8.164 & 11.720 & 8.758 & 9.778 \\
          $\mathbf{L}_{16}$ & {\textcolor{badresult}{\hspace{8pt}8.416}} & 4.159 & 13.624 & 1.989 & 11.084 & 6.558 & 5.445 & 9.192 & 8.436 & 8.996 \\
          $\mathbf{L}_{64}$ & 5.542 & 3.583 & 8.314 & 1.549 & 10.995 & {\textcolor{badresult}{\hspace{10pt}6.955}} & 4.140 & 8.174 & 8.299 & 8.944 \\
          $\mathbf{L}_{96}$ & 5.239 & 3.422 & 9.332 & 1.562 & 10.516 & {\textcolor{badresult}{\hspace{10pt}6.554}} & 3.498 & 7.072 & 8.700 & 9.703 \\
          $\mathbf{L}_{128}$ & 5.324 & 3.543 & 8.736 & 1.529 & 10.286 & 6.215 & 3.282 & 6.882 & 8.607 & 9.294 \\
          \bottomrule
      \end{tabularx}
\vspace{-5pt}
\caption{ \textbf{Ablations.}
Ablation results comparing reconstruction accuracy, cross-embodiment retargeting,
latent-space continuity, and interpolation smoothness. 
\textbf{Exp} denotes model variants: removing losses 
($-L_2$, $-L_2^{dist}$, $-L_2^{dir}$, or both), changing hidden sizes 
($H^b_a$), or changing latent dimension ($L_d$). 
Metrics include joint and tip RMSE for reconstruction; pinch- and random-motion
direction/distance errors for retargeting; joint/tip latent continuity; and mean
acceleration/jerk for interpolation.}
\vspace{-10pt}
\label{tab:latent_aba}
\end{table*}

\subsection{Ablation Results}
In this section, we are trying to answer the question about \textit{Effectiveness of the Latent Action Space:} (1) How well does the learned latent space function as a retargeting mechanism on its own? (2) What is the impact of different design choices within the latent space, as shown through ablation studies?

\noindent\textbf{Latent Replay Comparison.}
We further compare our latent action space against LAD~\cite{Bauer2025LAD}, a supervised latent-space retargeting method. 
To ensure a fair and challenging evaluation, we perform \emph{latent replay} by taking demonstrations from two embodiments and replaying them on the other two embodiments using each method's latent mapping. 
As shown in Table~\ref{tab:replay_latent_cmp}, our approach achieves a mean success rate of \textbf{0.82} and \textbf{0.81} on the two hand pairs (Ability+Inspire and Paxini+XHand), substantially outperforming LAD, which attains only \textbf{0.60} and \textbf{0.61}. 
This improvement is consistent across all tasks, with gains particularly pronounced on fine-grained manipulation tasks such as \textit{SC}, \textit{SoC}, and \textit{HB}, where LAD exhibits noticeable degradation. 
Notably, our method achieves these results without any supervision data or paired labels, relying solely on unsupervised latent alignment. 
These findings highlight that our latent space captures embodiment-invariant structure more effectively than supervised alternatives, enabling significantly more reliable cross-hand trajectory replay.

\begin{figure}[h]
    \centering
    \begin{subfigure}[b]{0.45\linewidth}
        \centering
        \includegraphics[width=\linewidth]{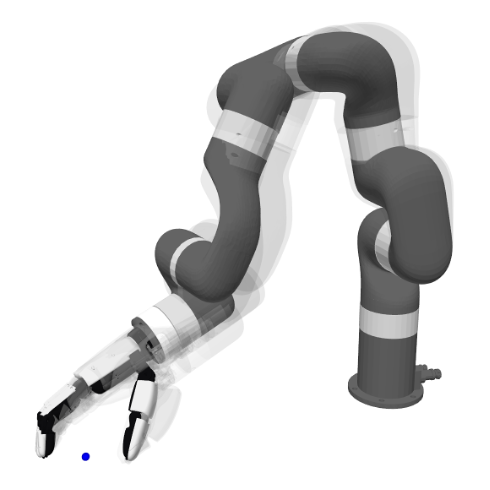}
        \caption{X-Hand}
        \label{fig:demo_xhand}
    \end{subfigure}
    \begin{subfigure}[b]{0.45\linewidth}
        \centering
        \includegraphics[width=\linewidth]{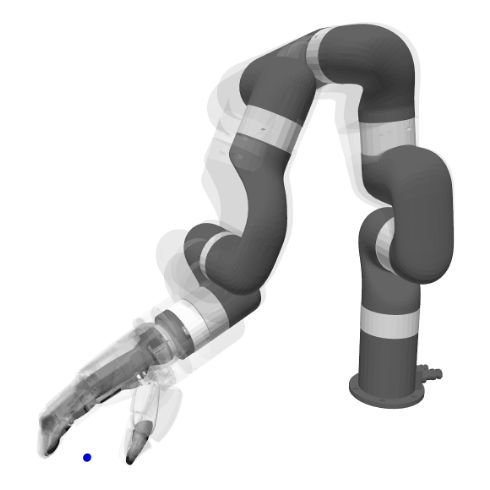}
        \caption{Inspire Hand}
        \label{fig:demo_inspire}
    \end{subfigure}
    \vspace{-8pt}
    \caption{\textbf{Latent Visualizations.} Latent decoding results cross embodiment.}
    \vspace{-8pt}
    \label{fig:demo_all_hands}
\end{figure}

\noindent
\textbf{Visual Result.} Figure~\ref{fig:demo_all_hands} shows latent decoding across different dexterous hands. We visualize one hand at full opacity and others with partial transparency, with the target grasp point marked in blue. Despite differing kinematics, all hands produce consistent poses from the same latent code, indicating that the learned latent space captures embodiment-invariant control.

\vspace{-8pt}

\noindent
\textbf{Design Choice Comparison.} 
We conduct a comprehensive ablation study to evaluate architectural and loss-design choices for the latent action space, summarized in Tab.~\ref{tab:latent_aba}. 
Our final configuration uses the $H_{64}^{128 \rightarrow 64}$ architecture with a latent dimension of 32. All metrics follow a ``lower is better'' convention, and the worst result compared with our method within each row are highlighted in green. Across reconstruction, cross-embodiment retargeting, latent continuity, and interpolation smoothness, our design choice achieves relatively stronger performance. Notably, performance remains stable across a wide range of architectures and latent dimensions, with degradation only occurring when the latent size is significantly increased (e.g., $L_{128}$), suggesting that excessively large latent spaces hinder embodiment-invariant structure. These results indicate that our chosen configuration offers an effective balance between model capacity and latent compactness.

And we explicitly write the evaluation metrics we design for this ablation study below:

\noindent
\textit{Recon Joint/Tip RMSE.}
Evaluates reconstruction fidelity of one hand. Synthetic random joint configurations are encoded and decoded, and we report the root-mean-square error (RMSE) between input and reconstructed joint angles (radians). In parallel, the original and reconstructed configurations are passed through each hand's forward-kinematics model to obtain fingertip positions, and we compute the RMSE of fingertip displacement (meters). Lower values indicate that the latent representation preserves hand configurations.

\noindent
\textit{Pinch and Random Tip Dir./Dist. Error}
Assesses cross-hand transfer for pinch and random grasps. Pinch poses and random poses are encoded on a source hand and decoded on each target hand; for each pose, we form a line from the thumb tip to the opposing fingertip. Directional error is measured as the angle between predicted and reference lines (degrees), and distance error is measured as the absolute difference in thumb--finger distance (meters). Smaller values for both components indicate less directional drift and more faithful preservation of pinch aperture across hands.

\noindent
\textit{Latent Continuity (Joint/Tip)}
Test the local smoothness of the latent manifold. Encoded hand latents are perturbed with isotropic Gaussian noise of standard deviation $\epsilon = 0.05$ and decoded back to joint angles, from which we compute the norm of the resulting joint-space deviation (radians). The corresponding finger tip effect is obtained by forwarding both perturbed and unperturbed reconstructions through forward kinematics and measuring the norm of fingertip displacement (meters). Small deviations indicate that the latent representation varies smoothly.

\noindent
\textit{Interp. Accel./Jerk Mean}
Characterizes the smoothness of latent-space interpolations. Two poses are encoded, and their latent codes are linearly interpolated. Decoding these intermediate codes yields fingertip trajectories, from which finite differences provide velocities and accelerations. Interp. Accel. Mean is the mean acceleration norm (meters per normalized interpolation step squared), while Interp. Jerk Mean is the mean norm of the jerk (the finite-difference derivative of acceleration, meters per step cubed). Lower values for both indicate smoother interpolation paths in the latent space.

\vspace{-8pt}
\section{Conclusion.} 
\vspace{-3pt}
In this work, we introduced XL-VLA, a vision–language–action framework equipped with a unified latent action space for scalable cross-embodiment dexterous manipulation. By learning an embodiment-invariant latent representation, our approach enables seamless training across diverse robotic hands and supports zero-shot generalization to new hand–task combinations. Extensive real-world experiments demonstrate that XL-VLA consistently outperforms standard VLA models and retargeting-based baselines, while offering a flexible and plug-and-play interface for newly introduced hands. Overall, our results highlight latent action spaces as a powerful foundation for building generalizable, data-efficient dexterous manipulation systems. We believe this work takes a step toward more unified and adaptable robotic manipulation frameworks capable of keeping pace with rapid hardware innovation.  
{
    \small
    \bibliographystyle{ieeenat_fullname}
    \bibliography{main}

@String(ICCV= {Int. Conf. Comput. Vis.})

@String(ICCV  = {ICCV})

@inproceedings{richardson2024isyhand,
  title   = {ISyHand: A Dexterous Multi-finger Robot Hand with an Articulated Palm},
  author  = {Richardson, Benjamin A and Gr{\"u}ninger, Felix and Mack, Lukas and Stueckler, Joerg and Kuchenbecker, Katherine J},
  booktitle = {Proceedings of the IEEE-RAS International Conference on Humanoid Robots (Humanoids)},
  address = {Seoul, Korea},
  year    = {2025}
}

@inproceedings{kuroda2024plexus,
  title   = {PLEXUS Hand: Lightweight Four-Motor Prosthetic Hand Enabling Precision-Lateral Dexterous Manipulation},
  author  = {Kuroda, Yuki and Takahashi, Tomoya and Beltran-Hernandez, Cristian C and Hamaya, Masashi and Tanaka, Kazutoshi},
  booktitle = {Proceedings of the IEEE International Conference on Rehabilitation Robotics (ICORR)},
  year    = {2025}
}

@article{sou2024moiretac,
  title={Moir\'{e}Tac: A Dual-Mode Visuotactile Sensor for Multidimensional Perception Using Moir\'{e} Pattern Amplification},
  author={Sou, Kit-Wa and Gong, Junhao and Li, Shoujie and Lyu, Chuqiao and Song, Ziwu and Mu, Shilong and Ding, Wenbo},
  journal={arXiv preprint arXiv:2509.12714},
  year={2025}
}

@article{xu2024multimodal,
  title={A multi-modal tactile fingertip design for robotic hands to enhance dexterous manipulation},
  author={Xu, Zhuowei and Si, Zilin and Zhang, Kevin and Kroemer, Oliver and Temel, Zeynep},
  journal={arXiv preprint arXiv:2510.05382},
  year={2025}
}

@article{liu2024dexndm,
  title={DexNDM: Closing the Reality Gap for Dexterous In-Hand Rotation via Joint-Wise Neural Dynamics Model},
  author={Liu, Xueyi and Wang, He and Yi, Li},
  journal={arXiv preprint arXiv:2510.08556},
  year={2025}
}

@article{hung2024avo,
  title={AVO: Amortized Value Optimization for Contact Mode Switching in Multi-Finger Manipulation},
  author={Hung, Adam and Yang, Fan and Kumar, Abhinav and Marinovic, Sergio Aguilera and Iba, Soshi and Zarrin, Rana Soltani and Berenson, Dmitry},
  journal={arXiv preprint arXiv:2510.07548},
  year={2025}
}

@inproceedings{chen2024exploiting,
  title={Exploiting Policy Idling for Dexterous Manipulation},
  author={Chen, Annie S and Brakel, Philemon and Bronars, Antonia and Xie, Annie and Huang, Sandy and Groth, Oliver and Bauza, Maria and Wulfmeier, Markus and Heess, Nicolas and Rao, Dushyant},
  booktitle={Proceedings of the IEEE/RSJ International Conference on Intelligent Robots and Systems (IROS)},
  year={2025}
}

@article{chi2024open,
  title={Open TeleDex: A Hardware-Agnostic Teleoperation System for Imitation Learning based Dexterous Manipulation},
  author={Chi, Xu and Zhang, Chao and Su, Yang and Dou, Lingfeng and Yang, Fujia and Zhao, Jiakuo and Zhou, Haoyu and Jia, Xiaoyou and Zhou, Yong and An, Shan},
  journal={arXiv preprint arXiv:2510.14771},
  year={2025}
}

@inproceedings{fang2024dexop,
  title={DEXOP: A Device for Robotic Transfer of Dexterous Human Manipulation},
  author={Fang, Hao-Shu and Romero, Branden and Xie, Yichen and Hu, Arthur and Huang, Bo-Ruei and Alvarez, Juan and Kim, Matthew and Margolis, Gabriel and Anbarasu, Kavya and Tomizuka, Masayoshi and others},
  booktitle={Workshop on Dexterous Manipulation at Robotics: Science and Systems (RSS)},
  note={Workshop paper},
  year={2025}
}

@inproceedings{xu2024dexumi,
  title={DexUMI: Using Human Hand as the Universal Manipulation Interface for Dexterous Manipulation},
  author={Xu, Mengda and Zhang, Han and Hou, Yifan and Xu, Zhenjia and Fan, Linxi and Veloso, Manuela and Song, Shuran},
  booktitle={Proceedings of the 9th Conference on Robot Learning (CoRL)},
  series   = {Proceedings of Machine Learning Research},
  volume   = {305},
  pages    = {437--459},
  year     = {2025}
}

@article{li2024scalable,
  title={Scalable Vision-Language-Action Model Pretraining for Robotic Manipulation with Real-Life Human Activity Videos},
  author={Li, Qixiu and Deng, Yu and Liang, Yaobo and Luo, Lin and Zhou, Lei and Yao, Chengtang and Zeng, Lingqi and Feng, Zhiyuan and Liang, Huizhi and Xu, Sicheng and others},
  journal={arXiv preprint arXiv:2510.21571},
  year={2025}
}

@article{qu2024eo,
  title={EO-1: Interleaved Vision-Text-Action Pretraining for General Robot Control},
  author={Qu, Delin and Song, Haoming and Chen, Qizhi and Chen, Zhaoqing and Gao, Xianqiang and Ye, Xinyi and Lv, Qi and Shi, Modi and Ren, Guanghui and Ruan, Cheng and others},
  journal={arXiv preprint arXiv:2508.21112},
  year={2025}
}

@inproceedings{yu2024forcevla,
  title={ForceVLA: Enhancing VLA Models with a Force-aware MoE for Contact-rich Manipulation},
  author={Yu, Jiawen and Liu, Hairuo and Yu, Qiaojun and Ren, Jieji and Hao, Ce and Ding, Haitong and Huang, Guangyu and Huang, Guofan and Song, Yan and Cai, Panpan and others},
  booktitle={Advances in Neural Information Processing Systems (NeurIPS)},
  note={Poster},
  year={2025}
}

@article{song2024ceed,
  title={CEED-VLA: Consistency Vision-Language-Action Model with Early-Exit Decoding},
  author={Song, Wenxuan and Chen, Jiayi and Ding, Pengxiang and Huang, Yuxin and Zhao, Han and Wang, Donglin and Li, Haoang},
  journal={arXiv preprint arXiv:2506.13725},
  year={2025}
}

@article{fei2024tro,
  title={T(R,O) Grasp: Efficient Graph Diffusion of Robot-Object Spatial Transformation for Cross-Embodiment Dexterous Grasping},
  author={Fei, Xin and Xu, Zhixuan and Fang, Huaicong and Zhang, Tianrui and Shao, Lin},
  journal={arXiv preprint arXiv:2510.12724},
  year={2025}
}

@inproceedings{puang2024pchands,
  title={PCHands: PCA-based Hand Pose Synergy Representation on Manipulators with N-DoF},
  author={Puang, En Yen and Ceola, Federico and Pasquale, Giulia and Natale, Lorenzo},
  booktitle={Proceedings of the IEEE-RAS International Conference on Humanoid Robots (Humanoids)},
  year={2025}
}

@article{wei2024omnidexgrasp,
  title={OmniDexGrasp: Generalizable Dexterous Grasping via Foundation Model and Force Feedback},
  author={Wei, Yi-Lin and Luo, Zhexi and Lin, Yuhao and Lin, Mu and Liang, Zhizhao and Chen, Shuoyu and Zheng, Wei-Shi},
  journal={arXiv preprint arXiv:2510.23119},
  year={2025}
}

@article{tan2025blm1,
  title        = {BLM$_1$: A Boundless Large Model for Cross-Space, Cross-Task, and Cross-Embodiment Learning},
  author       = {Tan, Wentao and Wang, Bowen and Zhi, Heng and Liu, Chenyu and Li, Zhe and Liu, Jian and Lin, Zengrong and Dai, Yukun and Chen, Yipeng and Yang, Wenjie and Xie, Enci and Xue, Hao and Ji, Baixu and Xu, Chen and Wang, Zhibin and Wang, Tianshi and Zhu, Lei and Shen, Heng Tao},
  journal      = {arXiv preprint arXiv:2510.24161},
  year         = {2025},
  url          = {https://arxiv.org/abs/2510.24161}
}

@article{zheng2025xvla,
  title        = {X-VLA: Soft-Prompted Transformer as Scalable Cross-Embodiment Vision-Language-Action Model},
  author       = {Zheng, Jinliang and Li, Jianxiong and Wang, Zhihao and Liu, Dongxiu and Kang, Xirui and Feng, Yuchun and Zheng, Yinan and Zou, Jiayin and Chen, Yilun and Zeng, Jia and Zhang, Ya-Qin and Pang, Jiangmiao and Liu, Jingjing and Wang, Tai and Zhan, Xianyuan},
  journal      = {arXiv preprint arXiv:2510.10274},
  year         = {2025},
  url          = {https://arxiv.org/abs/2510.10274}
}

@inproceedings{hou2025dita,
  title        = {Dita: Scaling Diffusion Transformer for Generalist Vision-Language-Action Policy},
  author       = {Hou, Zhi and Zhang, Tianyi and Xiong, Yuwen and Duan, Haonan and Pu, Hengjun and Tong, Ronglei and Zhao, Chengyang and Zhu, Xizhou and Qiao, Yu and Dai, Jifeng and Chen, Yuntao},
  booktitle    = {Proceedings of the IEEE/CVF International Conference on Computer Vision (ICCV)},
  year         = {2025}
}

@inproceedings{doshi2024crossformer,
  title        = {Scaling Cross-Embodied Learning: One Policy for Manipulation, Navigation, Locomotion and Aviation},
  author       = {Doshi, Ria and Walke, Homer and Mees, Oier and Dasari, Sudeep and Levine, Sergey},
  booktitle    = {Proceedings of the 8th Conference on Robot Learning (CoRL 2024)},
  series       = {Proceedings of Machine Learning Research},
  volume       = {270},
  pages        = {496--512},
  publisher    = {PMLR},
  year         = {2025},
  url          = {https://proceedings.mlr.press/v270/doshi25a.html}
}

@article{eftekhar2024ring,
  title        = {The One RING: a Robotic Indoor Navigation Generalist},
  author       = {Eftekhar, Ainaz and Hendrix, Rose and Weihs, Luca and Duan, Jiafei and Caglar, Ege and Salvador, Jordi and Herrasti, Alvaro and Han, Winson and VanderBil, Eli and Kembhavi, Aniruddha and Ehsani, Kiana and Zeng, Kuo-Hao and Krishna, Ranjay},
  journal      = {arXiv preprint arXiv:2412.14401},
  year         = {2024},
  url          = {https://arxiv.org/abs/2412.14401}
}

@inproceedings{yang2025multiloco,
  title        = {Multi-Loco: Unifying Multi-Embodiment Legged Locomotion via Reinforcement Learning Augmented Diffusion},
  author       = {Yang, Shunpeng and Fu, Zhen and Cao, Zhefeng and Junde, Guo and Wensing, Patrick and Zhang, Wei and Chen, Hua},
  booktitle    = {Proceedings of the 9th Conference on Robot Learning (CoRL 2025)},
  series       = {Proceedings of Machine Learning Research},
  volume       = {305},
  pages        = {1030--1048},
  publisher    = {PMLR},
  year         = {2025},
  url          = {https://proceedings.mlr.press/v305/}
}

@article{castro2025vamos,
  title        = {VAMOS: A Hierarchical Vision-Language-Action Model for Capability-Modulated and Steerable Navigation},
  author       = {Guaman Castro, Mateo and Rajagopal, Sidharth and Gorbatov, Daniel and Schmittle, Matt and Baijal, Rohan and Zhang, Octi and Scalise, Rosario and Talia, Sidharth and Romig, Emma and de Melo, Celso and Boots, Byron and Gupta, Abhishek},
  journal      = {arXiv preprint arXiv:2510.20818},
  year         = {2025},
  url          = {https://arxiv.org/abs/2510.20818}
}

@inproceedings{zakka2021xirl,
  title        = {XIRL: Cross-Embodiment Inverse Reinforcement Learning},
  author       = {Zakka, Kevin and Zeng, Andy and Florence, Pete and Tompson, Jonathan and Bohg, Jeannette and Dwibedi, Debidatta},
  booktitle    = {Proceedings of the 5th Conference on Robot Learning (CoRL 2021)},
  series       = {Proceedings of Machine Learning Research},
  volume       = {164},
  pages        = {537--546},
  publisher    = {PMLR},
  year         = {2021},
  url          = {https://proceedings.mlr.press/v164/}
}

@inproceedings{ren2025motiontracks,
  title        = {Motion Tracks: A Unified Representation for Human-Robot Transfer in Few-Shot Imitation Learning},
  author       = {Ren, Juntao and Sundaresan, Priya and Sadigh, Dorsa and Choudhury, Sanjiban and Bohg, Jeannette},
  booktitle    = {Proceedings of the IEEE International Conference on Robotics and Automation (ICRA)},
  year         = {2025}
}

@inproceedings{dan2025xsim,
  title        = {X-Sim: Cross-Embodiment Learning via Real-to-Sim-to-Real},
  author       = {Dan, Prithwish and Kedia, Kushal and Chao, Angela and Duan, Edward Weiyi and Pace, Maximus Adrian and Ma, Wei-Chiu and Choudhury, Sanjiban},
  booktitle    = {Proceedings of the 9th Conference on Robot Learning (CoRL 2025)},
  series       = {Proceedings of Machine Learning Research},
  volume       = {305},
  pages        = {816--833},
  publisher    = {PMLR},
  year         = {2025},
  url          = {https://proceedings.mlr.press/v305/}
}

@inproceedings{kim2025uniskill,
  title        = {UniSkill: Imitating Human Videos via Cross-Embodiment Skill Representations},
  author       = {Kim, Hanjung and Kang, Jaehyun and Kang, Hyolim and Cho, Meedeum and Kim, Seon Joo and Lee, Youngwoon},
  booktitle    = {Proceedings of the 9th Conference on Robot Learning (CoRL 2025)},
  series       = {Proceedings of Machine Learning Research},
  volume       = {305},
  publisher    = {PMLR},
  year         = {2025},
  url          = {https://proceedings.mlr.press/v305/}
}

@inproceedings{punamiya2025egobridge,
  title        = {EgoBridge: Domain Adaptation for Generalizable Imitation from Egocentric Human Data},
  author       = {Punamiya, Ryan and Patel, Dhruv and Aphiwetsa, Patcharapong and Kuppili, Pranav and Zhu, Lawrence Y. and Kareer, Simar and Hoffman, Judy and Xu, Danfei},
  booktitle    = {Advances in Neural Information Processing Systems (NeurIPS)},
  year         = {2025},
  note         = {Poster}
}

@inproceedings{niu2025human2locoman,
  title        = {Human2LocoMan: Learning Versatile Quadrupedal Manipulation with Human Pretraining},
  author       = {Niu, Yaru and Zhang, Yunzhe and Yu, Mingyang and Lin, Changyi and Li, Chenhao and Wang, Yikai and Yang, Yuxiang and Yu, Wenhao and Zhang, Tingnan and Li, Zhenzhen and Francis, Jonathan and Chen, Bingqing and Tan, Jie and Zhao, Ding},
  booktitle    = {Proceedings of Robotics: Science and Systems (RSS)},
  year         = {2025}
}

@article{seo2024legato,
  title        = {LEGATO: Cross-Embodiment Imitation Using a Grasping Tool},
  author       = {Seo, Mingyo and Park, H. Andy and Yuan, Shenli and Zhu, Yuke and Sentis, Luis},
  journal      = {arXiv preprint arXiv:2411.03682},
  year         = {2024},
  url          = {https://arxiv.org/abs/2411.03682}
}

@article{cao2025gdream,
  title        = {G-DReaM: Graph-conditioned Diffusion Retargeting across Multiple Embodiments},
  author       = {Cao, Zhefeng and Liu, Ben and Li, Sen and Zhang, Wei and Chen, Hua},
  journal      = {arXiv preprint arXiv:2505.20857},
  year         = {2025},
  url          = {https://arxiv.org/abs/2505.20857}
}

@article{liu2025trajbooster,
  title        = {TrajBooster: Boosting Humanoid Whole-Body Manipulation via Trajectory-Centric Learning},
  author       = {Liu, Jiacheng and Ding, Pengxiang and Zhou, Qihang and Wu, Yuxuan and Huang, Da and Peng, Zimian and Xiao, Wei and Zhang, Weinan and Yang, Lixin and Lu, Cewu and Wang, Donglin},
  journal      = {arXiv preprint arXiv:2509.11839},
  year         = {2025},
  url          = {https://arxiv.org/abs/2509.11839}
}

@article{wu2025cedex,
  title        = {CEDex: Cross-Embodiment Dexterous Grasp Generation at Scale from Human-like Contact Representations},
  author       = {Wu, Zhiyuan and Potamias, Rolandos Alexandros and Zhang, Xuyang and Zhang, Zhongqun and Deng, Jiankang and Luo, Shan},
  journal      = {arXiv preprint arXiv:2509.24661},
  year         = {2025},
  url          = {https://arxiv.org/abs/2509.24661}
}

@article{bauer2025latentdiff,
  title        = {Latent Action Diffusion for Cross-Embodiment Manipulation},
  author       = {Bauer, Erik and Nava, Elvis and Katzschmann, Robert K.},
  journal      = {arXiv preprint arXiv:2506.14608},
  year         = {2025},
  url          = {https://arxiv.org/abs/2506.14608}
}

@inproceedings{wen2025dexvla,
  title        = {DexVLA: Vision-Language Model with Plug-In Diffusion Expert for General Robot Control},
  author       = {Wen, Junjie and Zhu, Yichen and Li, Jinming and Tang, Zhibin and Shen, Chaomin and Feng, Feifei},
  booktitle    = {Proceedings of the 9th Conference on Robot Learning (CoRL 2025)},
  series       = {Proceedings of Machine Learning Research},
  volume       = {305},
  publisher    = {PMLR},
  year         = {2025},
  url          = {https://proceedings.mlr.press/v305/}
}

@article{bai2025roboswap,
  title        = {RoboSwap: A GAN-driven Video Diffusion Framework for Unsupervised Robot Arm Swapping},
  author       = {Bai, Yang and Yang, Liudi and Eskandar, George and Shen, Fengyi and Chen, Dong and Altillawi, Mohammad and Liu, Ziyuan and Kutyniok, Gitta},
  journal      = {arXiv preprint arXiv:2506.08632},
  year         = {2025},
  url          = {https://arxiv.org/abs/2506.08632}
}

@article{zhi2025flowaction,
  title        = {3DFlowAction: Learning Cross-Embodiment Manipulation from 3D Flow World Model},
  author       = {Zhi, Hongyan and Chen, Peihao and Zhou, Siyuan and Dong, Yubo and Wu, Quanxi and Han, Lei and Tan, Mingkui},
  journal      = {arXiv preprint arXiv:2506.06199},
  year         = {2025},
  url          = {https://arxiv.org/abs/2506.06199}
}

@article{zheng2025uniact,
  title        = {Universal Actions for Enhanced Embodied Foundation Models},
  author       = {Zheng, Jinliang and Li, Jianxiong and Liu, Dongxiu and Zheng, Yinan and Wang, Zhihao and Ou, Zhonghong and Liu, Yu and Liu, Jingjing and Zhang, Ya-Qin and Zhan, Xianyuan},
  journal      = {arXiv preprint arXiv:2501.10105},
  year         = {2025},
  url          = {https://arxiv.org/abs/2501.10105}
}

@article{zhang2025ate,
  title        = {Align-Then-stEer: Adapting the Vision-Language Action Models through Unified Latent Guidance},
  author       = {Zhang, Yang and Wang, Chenwei and Lu, Ouyang and Zhao, Yuan and Ge, Yunfei and Sun, Zhenglong and Li, Xiu and Zhang, Chi and Bai, Chenjia and Li, Xuelong},
  journal      = {arXiv preprint arXiv:2509.02055},
  year         = {2025},
  url          = {https://arxiv.org/abs/2509.02055}
}

@article{bu2025univla,
  title        = {UniVLA: Learning to Act Anywhere with Task-centric Latent Actions},
  author       = {Bu, Qingwen and Yang, Yanting and Yang, Jisong and Gao, Shenyuan and Ren, Guanghui and Yao, Maoqing and Luo, Ping and Li, Hongyang},
  journal      = {arXiv preprint arXiv:2505.06111},
  year         = {2025},
  url          = {https://arxiv.org/abs/2505.06111}
}

@article{wang2024latentalign,
  title        = {Cross-Embodiment Robot Manipulation Skill Transfer using Latent Space Alignment},
  author       = {Wang, Tianyu and Bhatt, Dwait and Wang, Xiaolong and Atanasov, Nikolay},
  journal      = {arXiv preprint arXiv:2406.01968},
  year         = {2024},
  url          = {https://arxiv.org/abs/2406.01968}
}

@article{davies2025tenma,
  title        = {Tenma: Robust Cross-Embodiment Robot Manipulation with Diffusion Transformer},
  author       = {Davies, Travis and Huang, Yiqi and Liu, Yunxin and Chen, Xiang and Liu, Huxian and Hu, Luhui},
  journal      = {arXiv preprint arXiv:2509.11865},
  year         = {2025},
  url          = {https://arxiv.org/abs/2509.11865}
}

@INPROCEEDINGS{BuQ-RSS-25,
  AUTHOR    = {Qingwen Bu AND Yanting Yang AND Jisong Cai AND Shenyuan Gao AND Guanghui Ren AND Maoqing Yao AND Ping Luo AND Hongyang Li},
  TITLE     = {{Learning to Act Anywhere with Task-centric Latent Actions}},
  BOOKTITLE = {Proceedings of Robotics: Science and Systems},
  YEAR      = {2025},
  ADDRESS   = {LosAngeles, CA, USA},
  MONTH     = {June},
  DOI       = {10.15607/RSS.2025.XXI.014},
  URL       = {https://www.roboticsproceedings.org/rss21/p014.pdf}
}

@article{Bauer2025LAD,
  title         = {Latent Action Diffusion for Cross-Embodiment Manipulation},
  author        = {Erik Bauer and Elvis Nava and Robert K. Katzschmann},
  journal       = {arXiv preprint arXiv:2506.14608},
  year          = {2025},
  eprint        = {2506.14608},
  archivePrefix = {arXiv},
  primaryClass  = {cs.RO},
  doi           = {10.48550/arXiv.2506.14608},
  url           = {https://arxiv.org/abs/2506.14608}
}

@article{Yang2025CoMo,
  title         = {CoMo: Learning Continuous Latent Motion from Internet Videos for Scalable Robot Learning},
  author        = {Jiange Yang and Yansong Shi and Haoyi Zhu and Mingyu Liu and Kaijing Ma and Yating Wang and Gangshan Wu and Tong He and Limin Wang},
  journal       = {arXiv preprint arXiv:2505.17006},
  year          = {2025},
  eprint        = {2505.17006},
  archivePrefix = {arXiv},
  primaryClass  = {cs.CV},
  doi           = {10.48550/arXiv.2505.17006},
  url           = {https://arxiv.org/abs/2505.17006}
}

@article{Dastider2025CycleVAEHBT,
  title         = {Cross-Embodiment Robotic Manipulation Synthesis via Guided Demonstrations through CycleVAE and Human Behavior Transformer},
  author        = {Apan Dastider and Hao Fang and Mingjie Lin},
  journal       = {arXiv preprint arXiv:2503.08622},
  year          = {2025},
  eprint        = {2503.08622},
  archivePrefix = {arXiv},
  primaryClass  = {cs.RO},
  doi           = {10.48550/arXiv.2503.08622},
  url           = {https://arxiv.org/abs/2503.08622}
}

@article{Wang2024LatentAlignment,
  title         = {Cross-Embodiment Robot Manipulation Skill Transfer using Latent Space Alignment},
  author        = {Tianyu Wang and Dwait Bhatt and Xiaolong Wang and Nikolay Atanasov},
  journal       = {arXiv preprint arXiv:2406.01968},
  year          = {2024},
  eprint        = {2406.01968},
  archivePrefix = {arXiv},
  primaryClass  = {cs.RO},
  doi           = {10.48550/arXiv.2406.01968},
  url           = {https://arxiv.org/abs/2406.01968}
}

@article{yin2025geort,
  title={Geometric Retargeting: A Principled, Ultrafast Neural Hand Retargeting Algorithm},
  author={Yin, Zhao-Heng and Wang, Changhao and Pineda, Luis and Bodduluri, Krishna and Wu, Tingfan and Abbeel, Pieter and Mukadam, Mustafa},
  journal={arXiv preprint arXiv:2503.07541},
  year={2025},
  url={http://arxiv.org/abs/2503.07541}
}

@article{lakshmipathy2024kinematic,
  title={Kinematic Motion Retargeting for Contact-Rich Anthropomorphic Manipulations},
  author={Lakshmipathy, Arjun S. and Hodgins, Jessica K. and Pollard, Nancy S.},
  journal={arXiv preprint arXiv:2402.04820},
  year={2024},
  url={http://arxiv.org/abs/2402.04820}
}

@article{xin2025analyzing,
  title={Analyzing Key Objectives in Human-to-Robot Retargeting for Dexterous Manipulation},
  author={Xin, Chendong and Yu, Mingrui and Jiang, Yongpeng and Zhang, Zhefeng and Li, Xiang},
  journal={arXiv preprint arXiv:2506.09384},
  year={2025},
  url={http://arxiv.org/abs/2506.09384}
}

@article{wen2025bytedexter,
  title={Dexterous Teleoperation of 20-DoF ByteDexter Hand via Human Motion Retargeting},
  author={Wen, Ruoshi and Zhang, Jiajun and Chen, Guangzeng and Cui, Zhongren and Du, Min and Gou, Yang and Han, Zhigang and Hu, Junkai and Huang, Liqun and Niu, Hao and Xu, Wei and Zhang, Haoxiang and Zhu, Zhengming and Li, Hang and Ren, Zeyu},
  journal={arXiv preprint arXiv:2507.03227},
  year={2025},
  url={http://arxiv.org/abs/2507.03227}
}

@article{chi2025openteledex,
  title={Open TeleDex: A Hardware-Agnostic Teleoperation System for Imitation Learning based Dexterous Manipulation},
  author={Chi, Xu and Zhang, Chao and Su, Yang and Dou, Lingfeng and Yang, Fujia and Zhao, Jiakuo and Zhou, Haoyu and Jia, Xiaoyou and Zhou, Yong and An, Shan},
  journal={arXiv preprint arXiv:2510.14771},
  year={2025},
  url={http://arxiv.org/abs/2510.14771}
}

@article{dong2025gex,
  title={GEX: Democratizing Dexterity with Fully-Actuated Dexterous Hand and Exoskeleton Glove},
  author={Dong, Yunlong and Liu, Xing and Wan, Jun and Deng, Zelin},
  journal={arXiv preprint arXiv:2506.04982},
  year={2025},
  url={http://arxiv.org/abs/2506.04982}
}

@article{lin2025dexflow,
  title={DexFlow: A Unified Approach for Dexterous Hand Pose Retargeting and Interaction},
  author={Lin, Xiaoyi and Yao, Kunpeng and Xu, Lixin and Wang, Xueqiang and Li, Xuetao and Wang, Yuchen and Li, Miao},
  journal={arXiv preprint arXiv:2505.01083},
  year={2025},
  url={http://arxiv.org/abs/2505.01083}
}

@article{mandi2025dexmachina,
  title={DexMachina: Functional Retargeting for Bimanual Dexterous Manipulation},
  author={Mandi, Zhao and Hou, Yifan and Fox, Dieter and Narang, Yashraj and Mandlekar, Ajay and Song, Shuran},
  journal={arXiv preprint arXiv:2505.24853},
  year={2025},
  url={http://arxiv.org/abs/2505.24853}
}

@article{xu2025dexplore,
  title={Dexplore: Scalable Neural Control for Dexterous Manipulation from Reference-Scoped Exploration},
  author={Xu, Sirui and Chao, Yu-Wei and Bian, Liuyu and Mousavian, Arsalan and Wang, Yu-Xiong and Gui, Liang-Yan and Yang, Wei},
  journal={arXiv preprint arXiv:2509.09671},
  year={2025},
  url={http://arxiv.org/abs/2509.09671}
}

@article{lin2025typetele,
  title={TypeTele: Releasing Dexterity in Teleoperation by Dexterous Manipulation Types},
  author={Lin, Yuhao and Wei, Yi-Lin and Liao, Haoran and Lin, Mu and Xing, Chengyi and Li, Hao and Zhang, Dandan and Cutkosky, Mark and Zheng, Wei-Shi},
  journal={arXiv preprint arXiv:2507.01857},
  year={2025},
  url={http://arxiv.org/abs/2507.01857}
}

@article{brohan2023rt2visionlanguageactionmodelstransfer,
  title={RT-2: Vision-language-action models transfer web knowledge to robotic control},
  author={Brohan, Anthony and Brown, Noah and Carbajal, Justice and Chebotar, Yevgen and Chen, Xi and Choromanski, Krzysztof and Ding, Tianli and Driess, Danny and Dubey, Avinava and Finn, Chelsea and others},
  journal={arXiv preprint arXiv:2307.15818},
  year={2023}
}

@article{kim2024openvla,
  title={OpenVLA: An open-source vision-language-action model},
  author={Kim, Moo Jin and Pertsch, Karl and Karamcheti, Siddharth and Xiao, Ted and Balakrishna, Ashwin and Nair, Suraj and Rafailov, Rafael and Foster, Ethan and Lam, Grace and Sanketi, Pannag and others},
  journal={arXiv preprint arXiv:2406.09246},
  year={2024}
}

@article{wen2024tinyvlafastdataefficientvisionlanguageaction,
  title={TinyVLA: Towards fast, data-efficient vision-language-action models for robotic manipulation},
  author={Wen, Junjie and Zhu, Yichen and Li, Jinming and Zhu, Minjie and Wu, Kun and Xu, Zhiyuan and Liu, Ning and Cheng, Ran and Shen, Chaomin and Peng, Yaxin and others},
  journal={arXiv preprint arXiv:2409.12514},
  year={2024}
}

@article{zhao2023learning,
  title={Learning fine-grained bimanual manipulation with low-cost hardware},
  author={Zhao, Tony Z and Kumar, Vikash and Levine, Sergey and Finn, Chelsea},
  journal={arXiv preprint arXiv:2304.13705},
  year={2023}
}

@article{beyer2024paligemma,
  title={PaliGemma: A versatile 3B VLM for transfer},
  author={Beyer, Lucas and Steiner, Andreas and Pinto, Andr{\'e} Susano and Kolesnikov, Alexander and Wang, Xiao and Salz, Daniel and Neumann, Maxim and Alabdulmohsin, Ibrahim and Tschannen, Michael and Bugliarello, Emanuele and others},
  journal={arXiv preprint arXiv:2407.07726},
  year={2024}
}

@article{bunny-visionpro,
    title   = {Bunny-VisionPro: Real-Time Bimanual Dexterous Teleoperation for Imitation Learning}, 
    author  = {Runyu Ding and Yuzhe Qin and Jiyue Zhu and Chengzhe Jia and Shiqi Yang and Ruihan Yang and Xiaojuan Qi and Xiaolong Wang},
    year    = {2024},
    url     = {https://arxiv.org/abs/2407.03162}, 
}

@misc{pi0,
      title={$\pi_0$: A Vision-Language-Action Flow Model for General Robot Control}, 
      author={Kevin Black and Noah Brown and Danny Driess and Adnan Esmail and Michael Equi and Chelsea Finn and Niccolo Fusai and Lachy Groom and Karol Hausman and Brian Ichter and Szymon Jakubczak and Tim Jones and Liyiming Ke and Sergey Levine and Adrian Li-Bell and Mohith Mothukuri and Suraj Nair and Karl Pertsch and Lucy Xiaoyang Shi and James Tanner and Quan Vuong and Anna Walling and Haohuan Wang and Ury Zhilinsky},
      year={2024},
      eprint={2410.24164},
      archivePrefix={arXiv},
      primaryClass={cs.LG},
      url={https://arxiv.org/abs/2410.24164}, 
}

@article{dong2025-dtc,
  title={Digital Twin Catalog: A Large-Scale Photorealistic 3D Object Digital Twin Dataset},
  author={Dong, Zhao and Chen, Ka and Lv, Zhaoyang and Yu, Hong-Xing and Zhang, Yunzhi and Zhang, Cheng and Zhu, Yufeng and Tian, Stephen and Li, Zhengqin and Moffatt, Geordie and others},
  journal={arXiv preprint arXiv:2504.08541},
  year={2025}
}

@article{calli2017yale,
  title={Yale-CMU-Berkeley dataset for robotic manipulation research},
  author={Calli, Berk and Singh, Arjun and Bruce, James and Walsman, Aaron and Konolige, Kurt and Srinivasa, Siddhartha and Abbeel, Pieter and Dollar, Aaron M},
  journal={The International Journal of Robotics Research},
  volume={36},
  number={3},
  pages={261--268},
  year={2017},
  publisher={SAGE Publications Sage UK: London, England}
}

@article{jiang2025gsworld,
  title={Gsworld: Closed-loop photo-realistic simulation suite for robotic manipulation},
  author={Jiang, Guangqi and Chang, Haoran and Qiu, Ri-Zhao and Liang, Yutong and Ji, Mazeyu and Zhu, Jiyue and Dong, Zhao and Zou, Xueyan and Wang, Xiaolong},
  journal={arXiv preprint arXiv:2510.20813},
  year={2025}
}

@article{jiang2024robots,
  title={Robots pre-train robots: Manipulation-centric robotic representation from large-scale robot datasets},
  author={Jiang, Guangqi and Sun, Yifei and Huang, Tao and Li, Huanyu and Liang, Yongyuan and Xu, Huazhe},
  journal={arXiv preprint arXiv:2410.22325},
  year={2024}
}

@article{ye2025power,
  title={From Power to Precision: Learning Fine-grained Dexterity for Multi-fingered Robotic Hands},
  author={Ye, Jianglong and Wei, Lai and Jiang, Guangqi and Jing, Changwei and Zou, Xueyan and Wang, Xiaolong},
  journal={arXiv preprint arXiv:2511.13710},
  year={2025}
}

@inproceedings{huang2024diffusion,
  title={Diffusion reward: Learning rewards via conditional video diffusion},
  author={Huang, Tao and Jiang, Guangqi and Ze, Yanjie and Xu, Huazhe},
  booktitle={European Conference on Computer Vision},
  pages={478--495},
  year={2024},
  organization={Springer}
}

@article{pan2024roboduet,
  title={Roboduet: A framework affording mobile-manipulation and cross-embodiment},
  author={Pan, Guoping and Ben, Qingwei and Yuan, Zhecheng and Jiang, Guangqi and Ji, Yandong and Pang, Jiangmiao and Liu, Houde and Xu, Huazhe},
  journal={arXiv preprint arXiv:2403.17367},
  volume={6},
  year={2024}
}

@article{jing2026contact,
  title={Contact-Aware Neural Dynamics},
  author={Jing, Changwei and Bandi, Jai Krishna and Ye, Jianglong and Duan, Yan and Abbeel, Pieter and Wang, Xiaolong and Yi, Sha},
  journal={arXiv preprint arXiv:2601.12796},
  year={2026}
}

@ticle{yan2025ace,
  title={ACE-f: A cross embodiment foldable system with force feedback for dexterous teleoperation},
  author={Yan, Rui and Fu, Jiajian and Yang, Shiqi and Paulsen, Lars and Cheng, Xuxin and Wang, Xiaolong},
  journal={arXiv preprint arXiv:2511.20887},
  year={2025}
}

@article{ben2025homie,
  title={Homie: Humanoid loco-manipulation with isomorphic exoskeleton cockpit},
  author={Ben, Qingwei and Jia, Feiyu and Zeng, Jia and Dong, Junting and Lin, Dahua and Pang, Jiangmiao},
  journal={arXiv preprint arXiv:2502.13013},
  year={2025}
}
}

\newpage

\twocolumn[
    \centering
    {\Large \bfseries Cross-Hand Latent Representation for Vision-Language-Action Models\par}
    \vspace{0.5em}

    \vspace{1em}
]

\section{Appendix}

\subsection{Latent Visualizations}
\begin{figure}[h]
    \centering
    \begin{subfigure}[b]{0.45\linewidth}
        \centering
        \includegraphics[width=\linewidth]{imgs/demo_transparent_1.png}
        \caption{X-Hand}
        \label{fig:demo_xhand}
    \end{subfigure}
    \begin{subfigure}[b]{0.45\linewidth}
        \centering
        \includegraphics[width=\linewidth]{imgs/demo_transparent_2.png}
        \caption{Inspire Hand}
        \label{fig:demo_inspire}
    \end{subfigure}
    \begin{subfigure}[b]{0.45\linewidth}
        \centering
        \includegraphics[width=\linewidth]{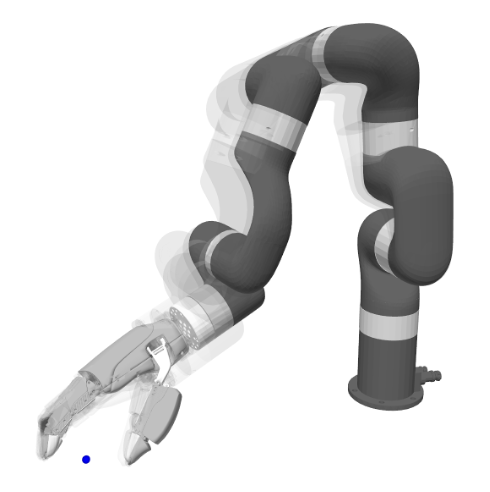}
        \caption{Paxini Hand}
        \label{fig:demo_paxini}
    \end{subfigure}
    \begin{subfigure}[b]{0.45\linewidth}
        \centering
        \includegraphics[width=\linewidth]{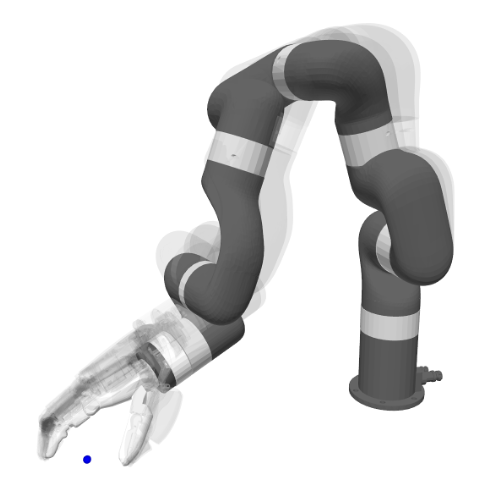}
        \caption{Ability Hand}
        \label{fig:demo_ability}
    \end{subfigure}
    \vspace{-8pt}
    \caption{\textbf{More Latent Visualizations.} Latent decoding results cross embodiment.}
    \vspace{-8pt}
    \label{fig:supp_latent}
\end{figure}

In addition to Figure 5, we include visualizations of additional hands in Figure~\ref{fig:supp_latent}. This figure illustrates how the same latent representation is decoded across all four hands featured in our main paper. Furthermore, Figure~\ref{fig:latent_traj} presents a continuous trajectory rendered for all hands, with the X-Hand highlighted for clarity.

\subsection{Hardware Setup}
\begin{figure}[h]
    \centering
    \includegraphics[width=\linewidth]{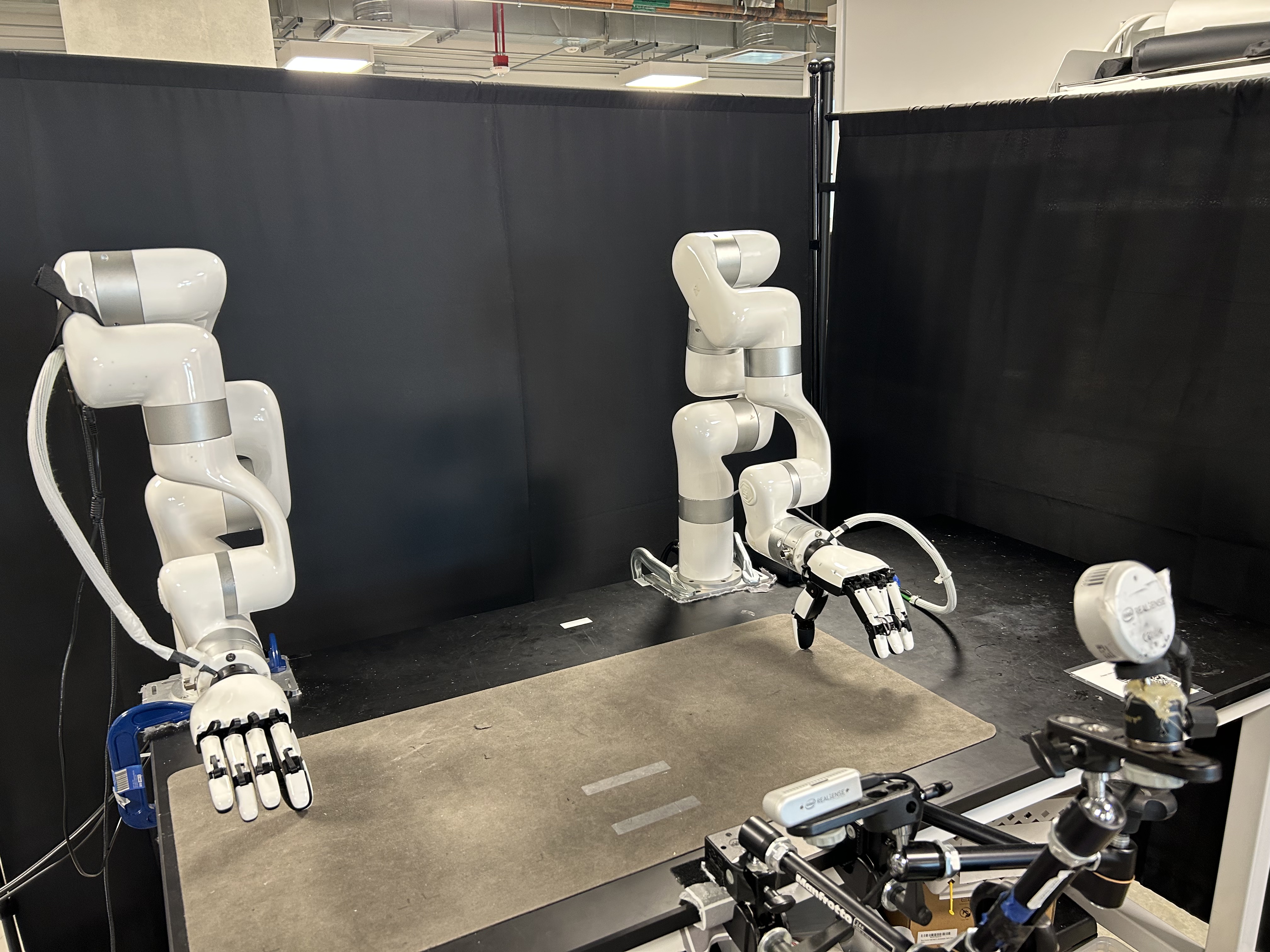}
    \caption{\textbf{xArm Camera Setup.} We use a single RealSense L515 camera with the front view. Note that the D435 camera here is not used for \ourmethod{}.}
    \label{fig:cam_setup}
\end{figure}

\begin{figure}[h]
    \centering
    \includegraphics[width=\linewidth]{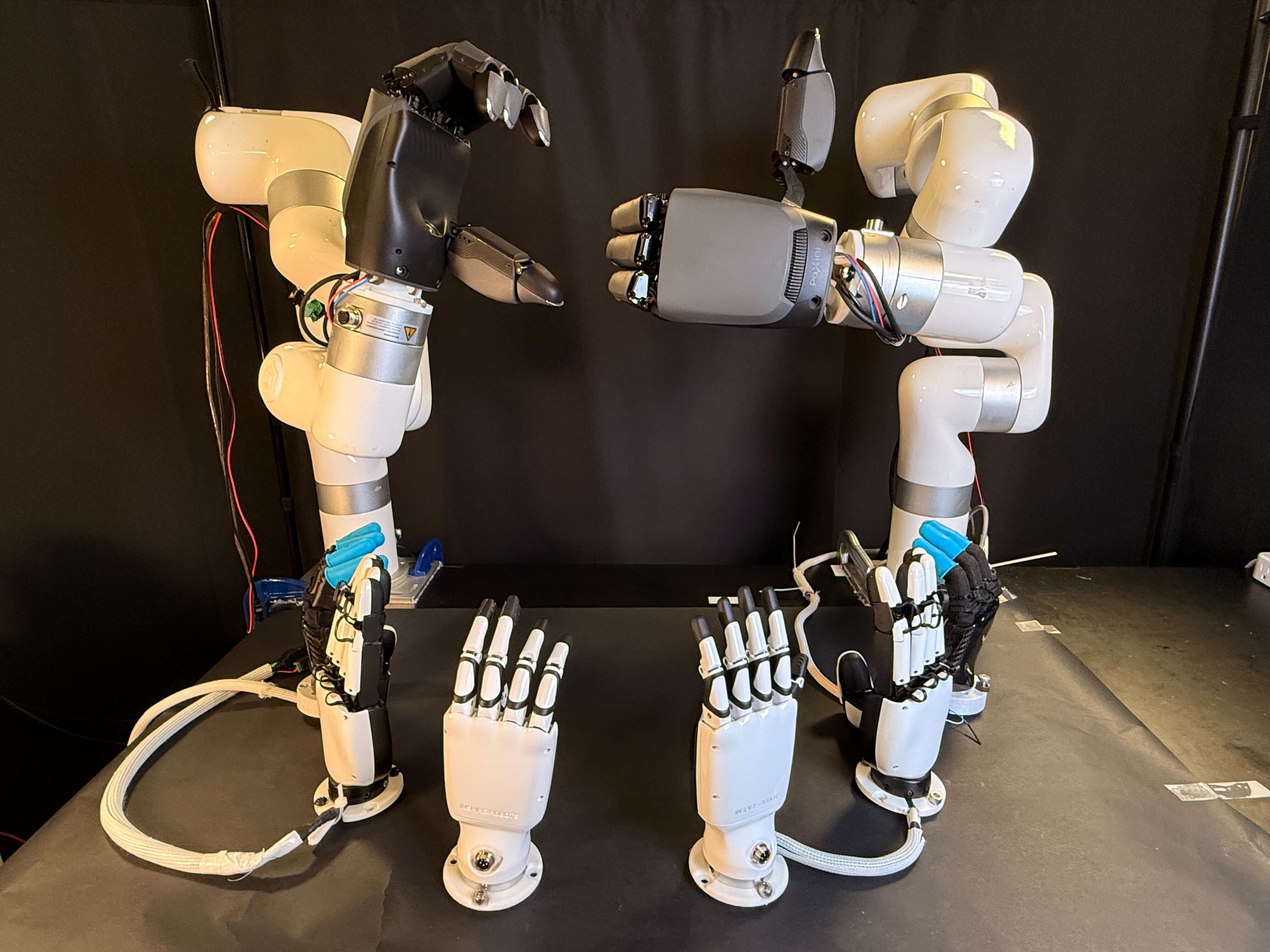}
    \caption{\textbf{Dexterous Hands.} We use 4 kinds of hands, with various shapes, scales, degrees of freedom, and actuated joints.}
    \label{fig:dex_hands}
\end{figure}

\noindent\textbf{Tabletop Scene Description.} For the real-world experiments, we use a bimanual arm with tabletop settings. The arms are mounted on the edge of the table. The distance between the arms are 80cm. Each hand is connected to the end effector of the arm with 3D-printed mounts. Figure~\ref{fig:cam_setup} shows the real-world tabletop scene with a pair of XHand, and figure~\ref{fig:dex_hands} includes all the dexterous hands we use in our experiments.

\begin{figure}[h]
    \centering
    \includegraphics[width=\linewidth]{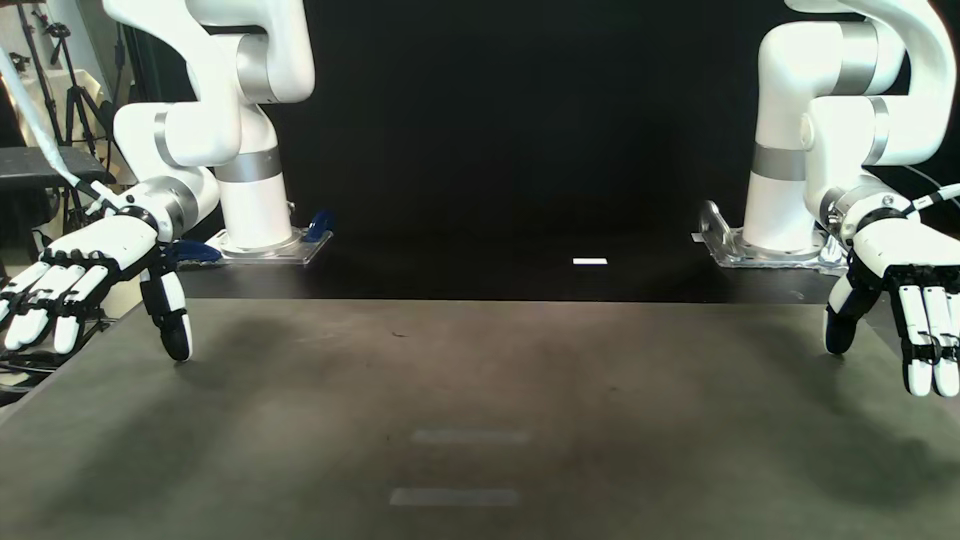}
    \caption{\textbf{xArm Camera View.} This is what our camera sees and also the input for \ourmethod{} and all the baseline methods.}
    \label{fig:cam_view}
\end{figure}

\noindent\textbf{Camera Description.}
We use a single RealSense L515 camera (round-shaped) mounted in front of the bimanual arms as the input view of the policy training. The camera pose is shown in figure~\ref{fig:cam_setup} and the camera view is in figure~\ref{fig:cam_view}. The raw resolution of RGB recordings from the camera is 960 $\times$ 540.

\begin{figure}[h]
    \centering
    \includegraphics[width=\linewidth]{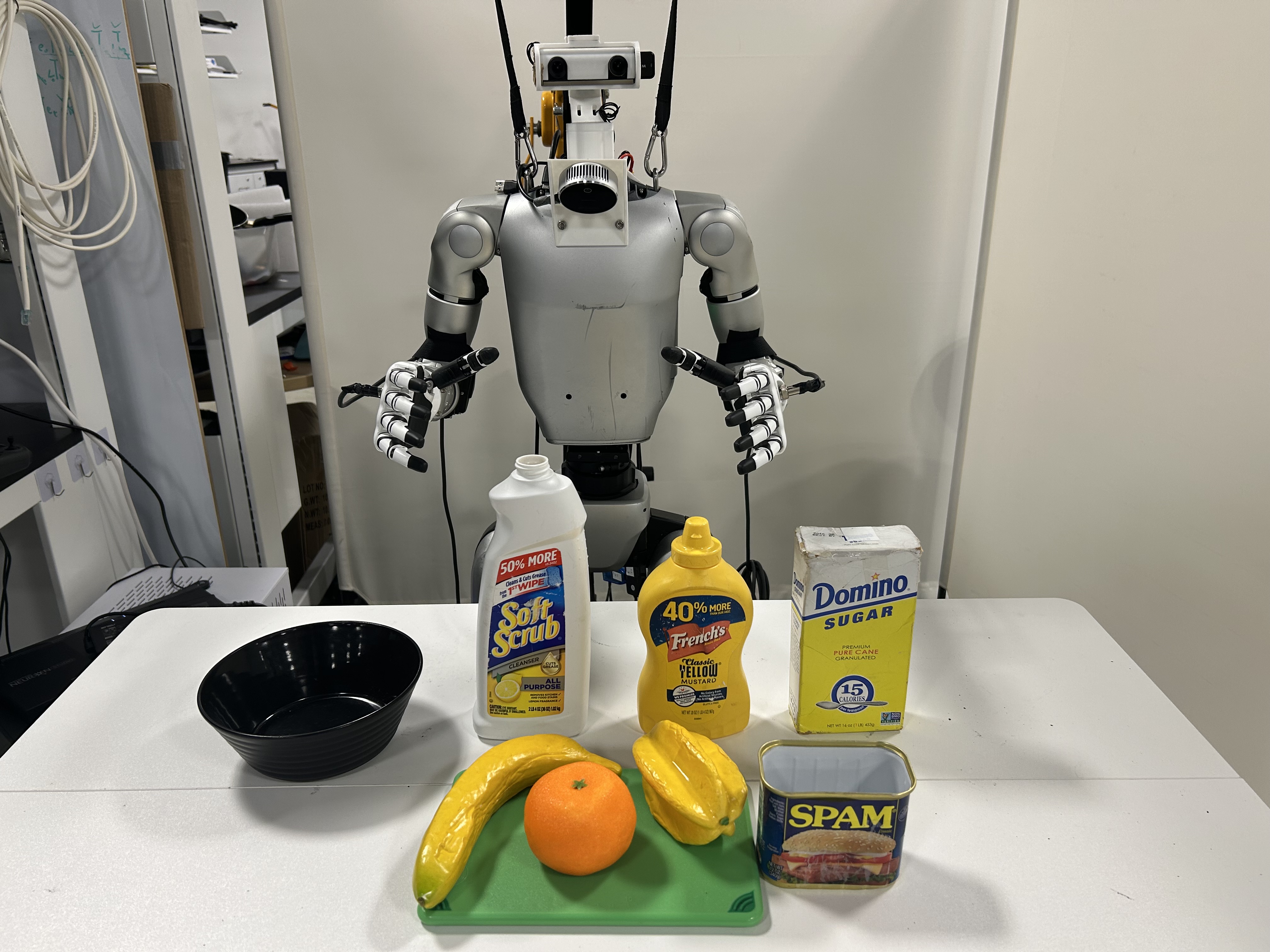}
    \caption{\textbf{G1 Scene.} We mount an L515 camera near the neck of G1 to have an egocentric view.}
    \label{fig:g1-scene}
\end{figure}

\begin{figure}[h]
    \centering
    \includegraphics[width=\linewidth]{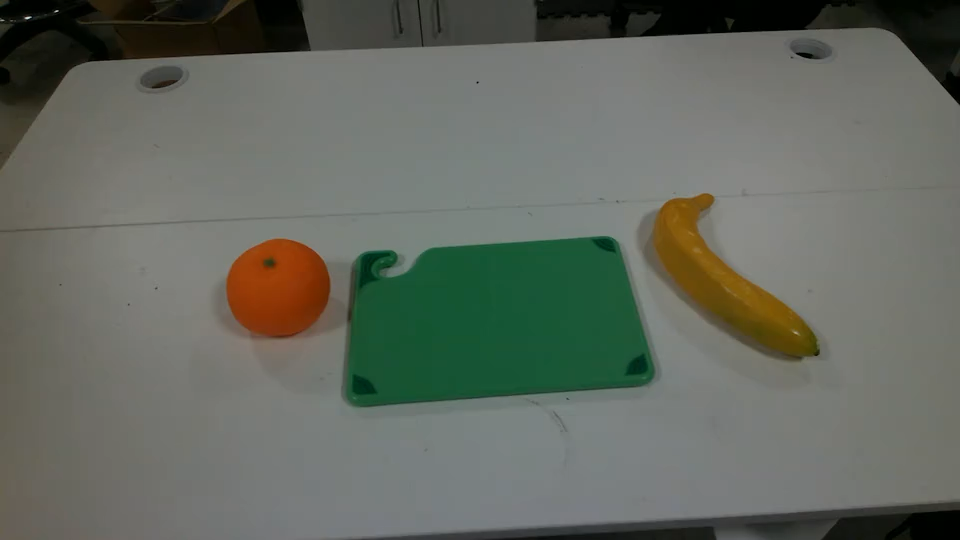}
    \caption{\textbf{G1 Egocentric Camera View.}}
    \label{fig:g1-cam-view}
\end{figure}

\noindent\textbf{Humanoid Scene Description.} Similar to xArm, we let G1 stand in front of a table. We use the same camera setting and mount L515 on the chest of G1 to have an egocentric view. Consider the mechanic design of G1, we only use the Inspire hand since it is light. Figure~\ref{fig:g1-scene} and ~\ref{fig:g1-cam-view} show the real-world G1 setting.

\begin{figure}[h]
    \centering
    \includegraphics[width=\linewidth]{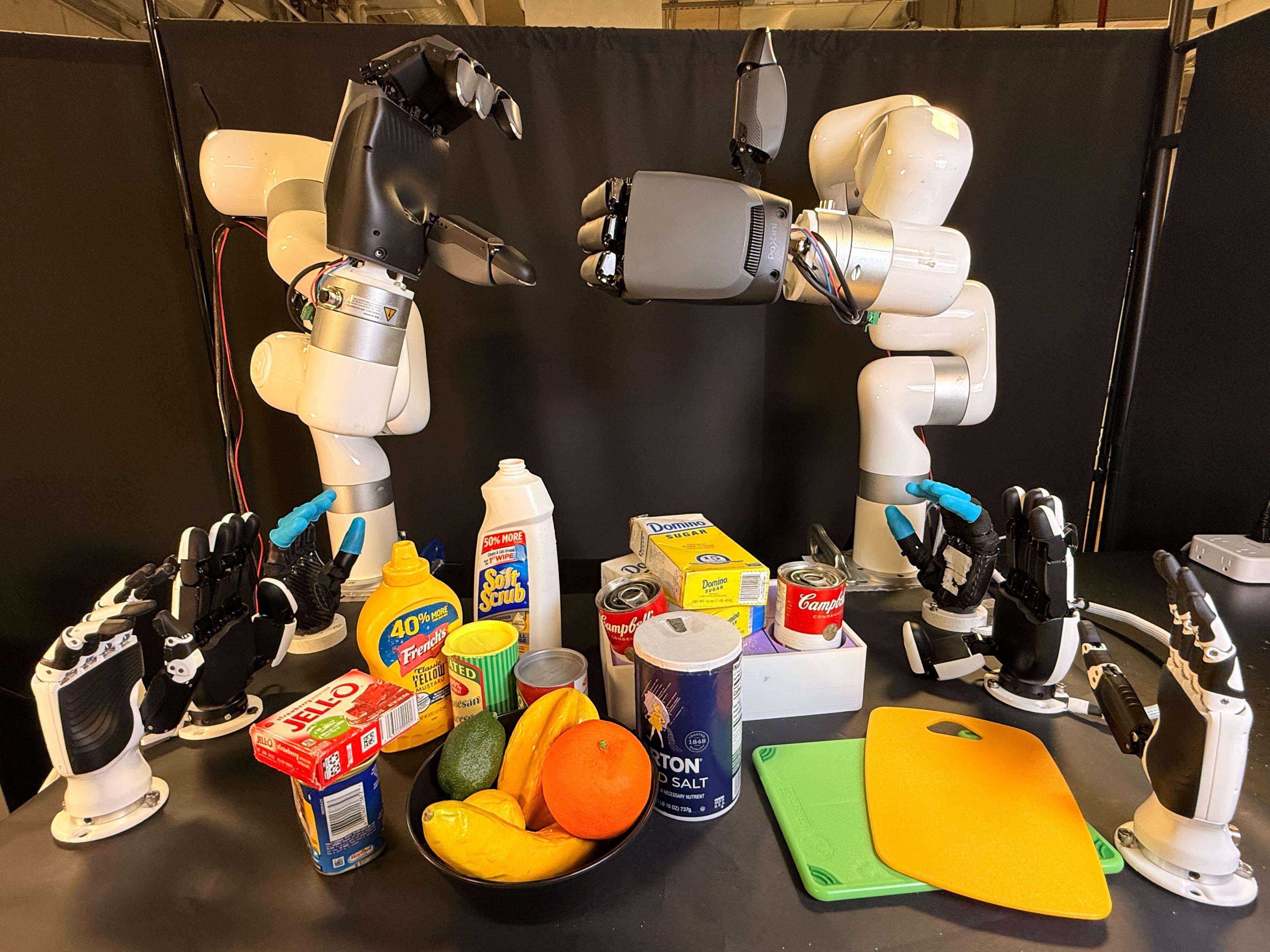}
    \caption{\textbf{Objects.} We use various everyday objects from existing datasets. They vary in scale, shape, texture, weight, etc., thus requiring the manipulation policy to be robust.}
    \label{fig:obj}
\end{figure}

\noindent\textbf{Object Description.}
To demonstrate that \ourmethod{} is capable of doing various manipulation tasks, we use diverse objects in our experiments, most of which are common everyday objects from existing datasets~\cite{dong2025-dtc, calli2017yale}. The objects vary in scale, shape, texture, weight, etc. All the objects used in listed in figure~\ref{fig:obj}.

\subsection{Policy Learning Details}

\begin{figure}[h]
    \centering
    \includegraphics[width=\linewidth]{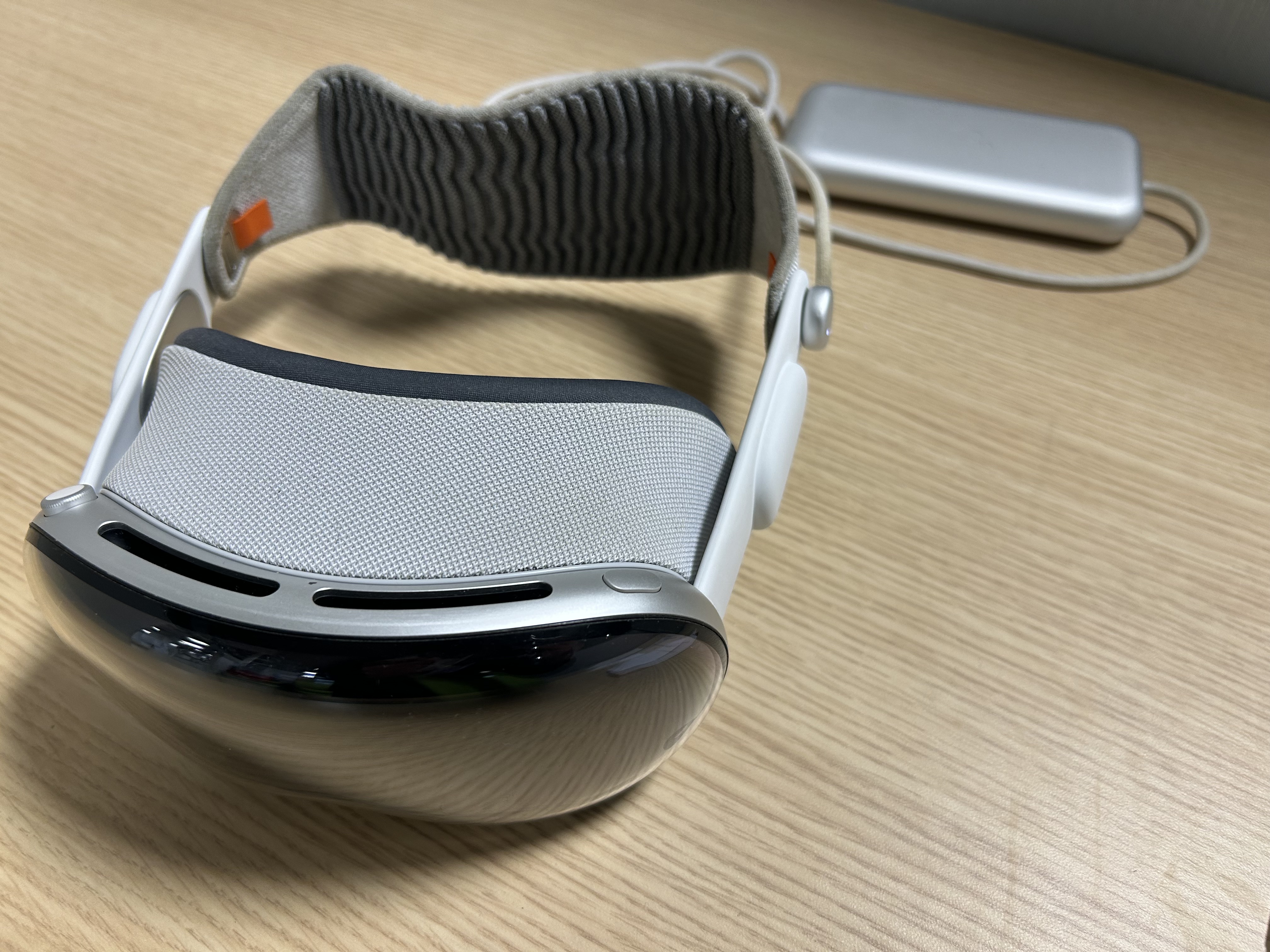}
    \caption{\textbf{Apple Vision Por for Data Collection.} Apple Vision Pro is used to track the human teleoperator's hands and wrists. Then the tracked data is processed with retargeting and Inverse Kinematics to control the real-world robots.}
    \label{fig:avp}
\end{figure}

\noindent\textbf{xArm Data Collection.}
We use Apple Vision Pro as the data collection tool~\cite{bunny-visionpro}, shown in figure~\ref{fig:avp}. During data collection, our teleoperator wears the VR headset and get the tracked hand poses and wrist poses from the headset. We use these data to do robot hand retargeting and Inverse Kinematics.

\noindent\textbf{Unitree G1 Data Collection.}
G1 is standing in front of a table, and we adapt HOMIE~\cite{ben2025homie} and ACE-F~\cite{yan2025ace} to do the upper-body teleoperation. We only use the upper-body system from HOMIE and replace their glove with the MANUS Mocap glove.

\begin{figure}[h]
    \centering
    \includegraphics[height=0.33\textheight, angle=-90]{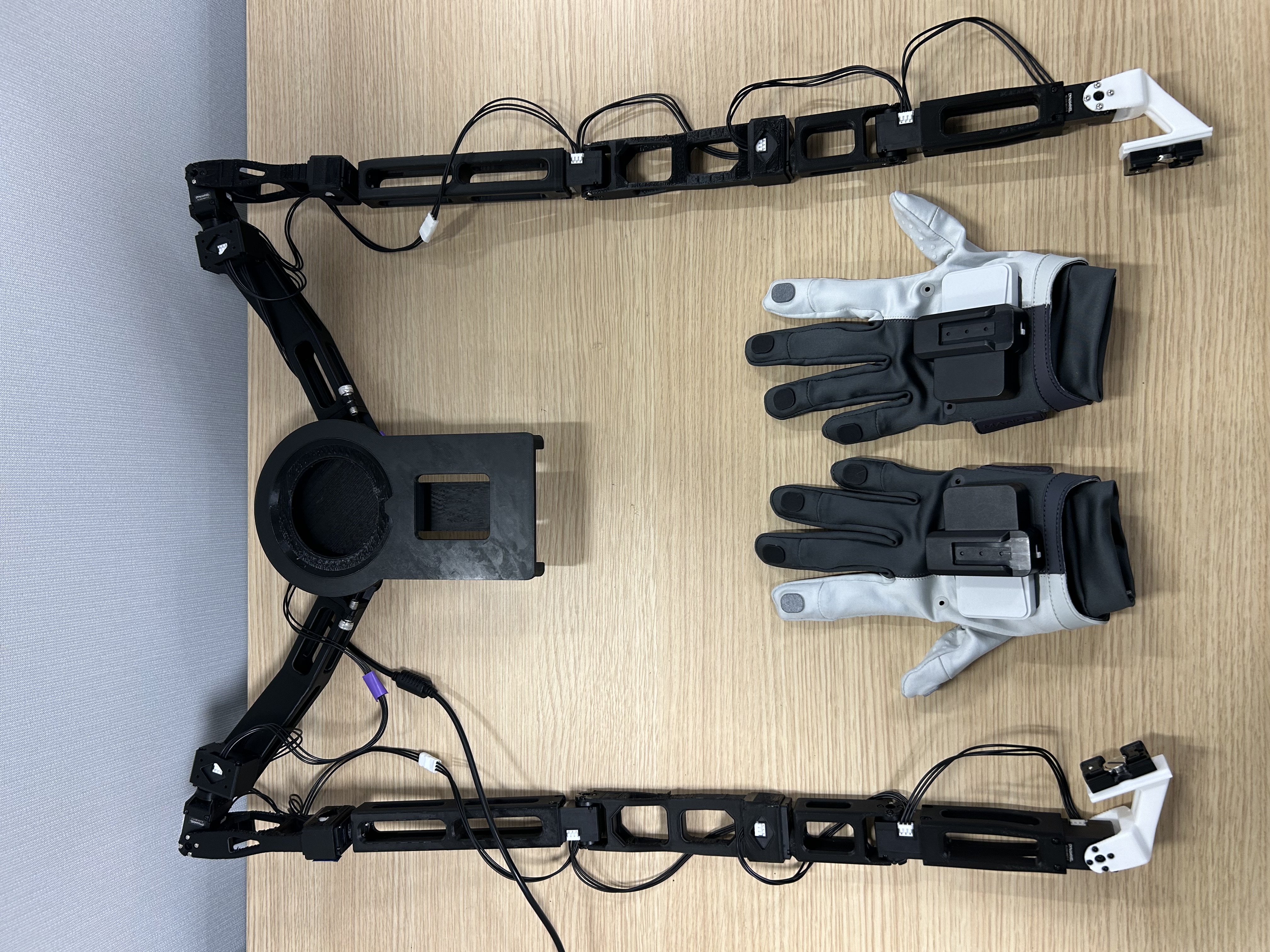}
    \caption{\textbf{G1 Teleoperation System.} We build the G1 upper-body teleoperation system from HOMIE~\cite{ben2025homie}. We use a pair of MANUS Mocap glove to track the human hand pose.}
    \label{fig:g1-teleop}
\end{figure}

\begin{figure}[h]
    \centering
    \begin{tabular}{cc}
        \begin{subfigure}{0.48\linewidth}
            \centering
            \includegraphics[width=\linewidth]{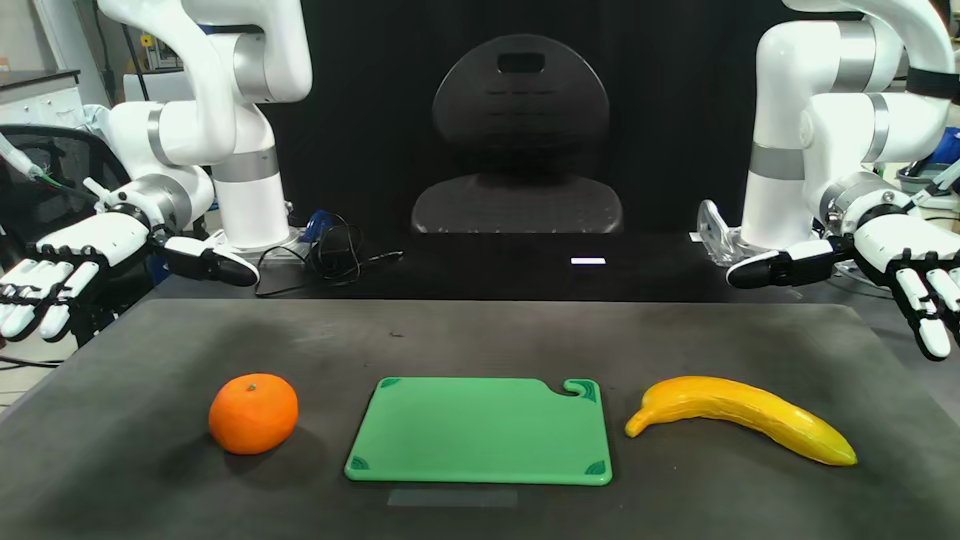}
        \end{subfigure} &
        \begin{subfigure}{0.48\linewidth}
            \centering
            \includegraphics[width=\linewidth]{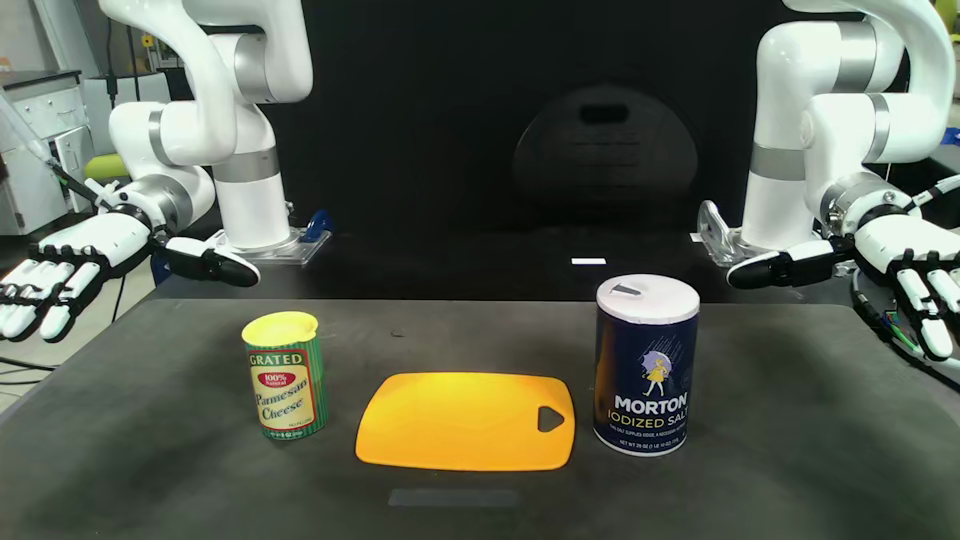}
        \end{subfigure} \\[6pt]

        \begin{subfigure}{0.48\linewidth}
            \centering
            \includegraphics[width=\linewidth]{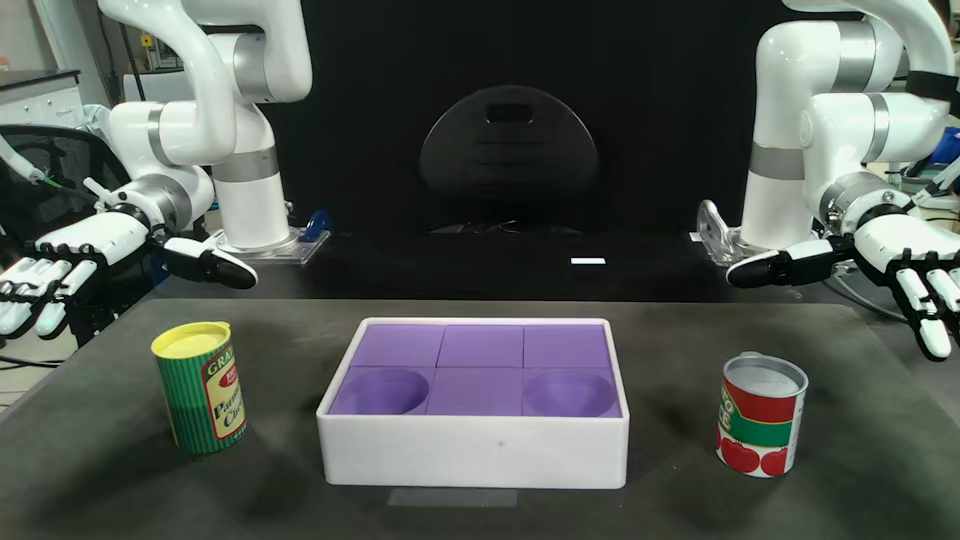}
        \end{subfigure} &
        \begin{subfigure}{0.48\linewidth}
            \centering
            \includegraphics[width=\linewidth]{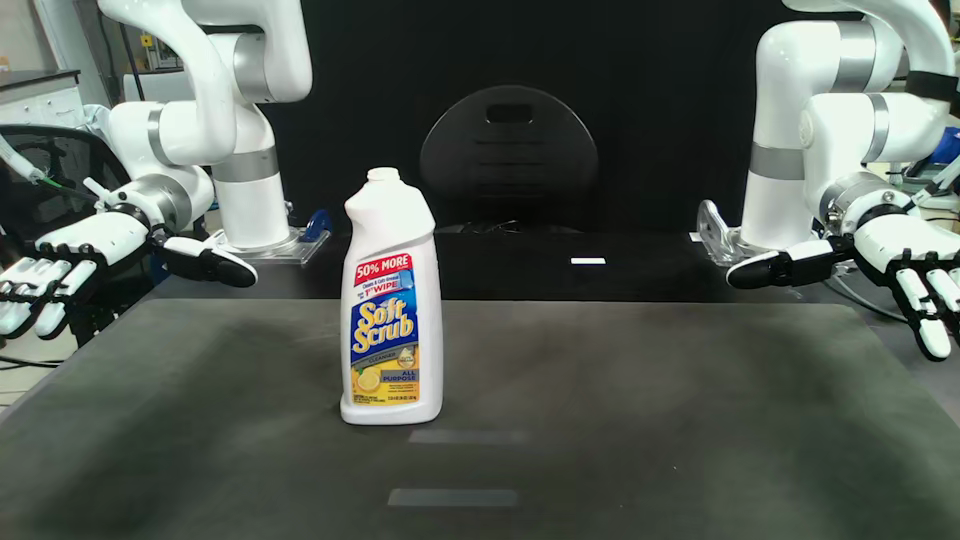}
        \end{subfigure} \\[6pt]

        \begin{subfigure}{0.48\linewidth}
            \centering
            \includegraphics[width=\linewidth]{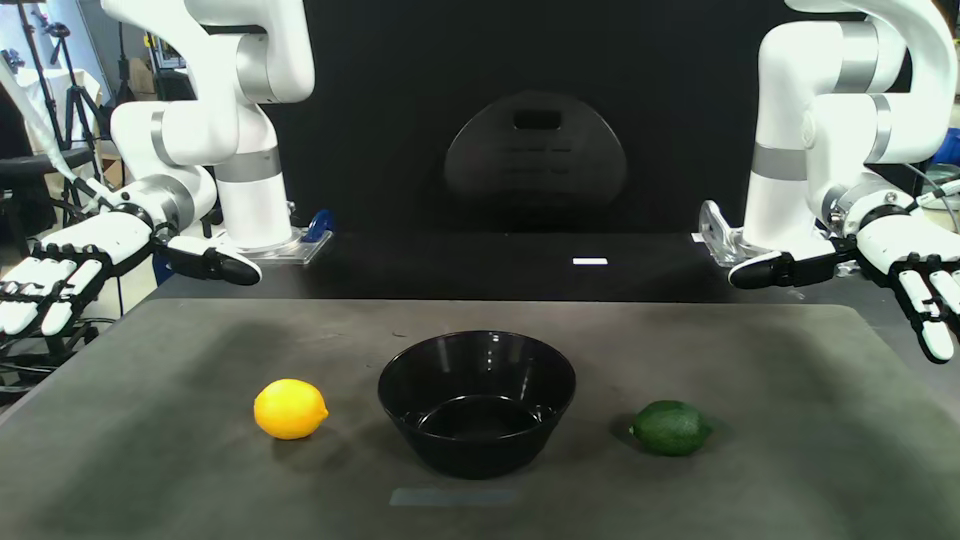}
        \end{subfigure} &
        \begin{subfigure}{0.48\linewidth}
            \centering
            \includegraphics[width=\linewidth]{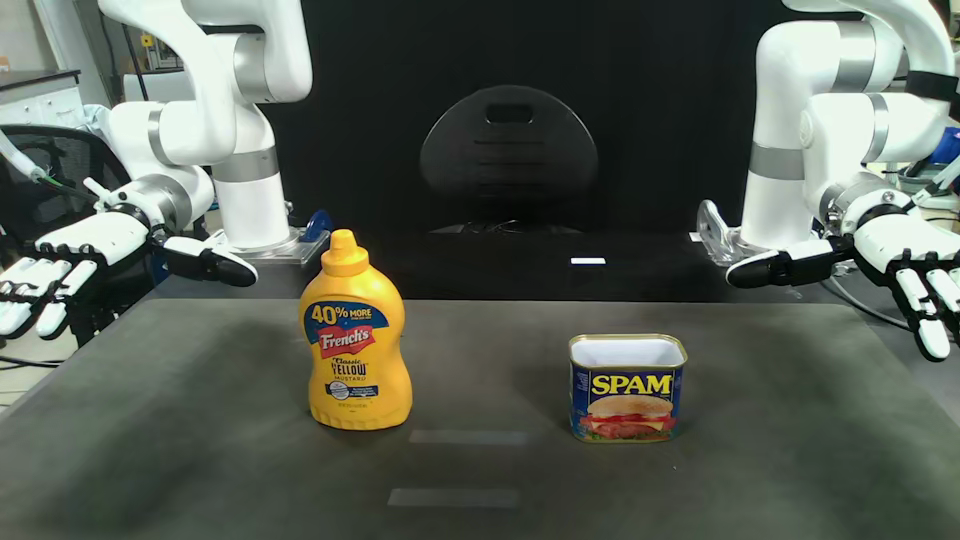}
        \end{subfigure} \\[6pt]

        \begin{subfigure}{0.48\linewidth}
            \centering
            \includegraphics[width=\linewidth]{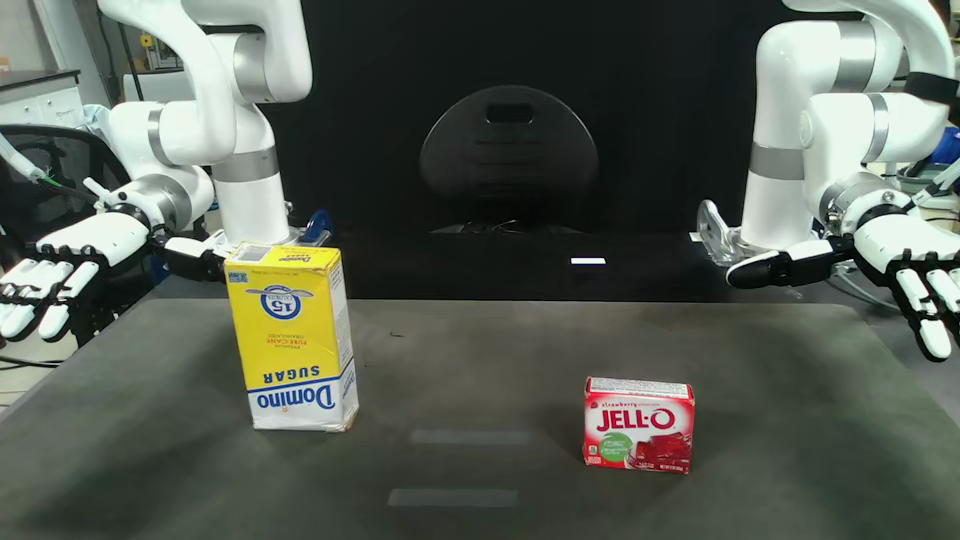}
        \end{subfigure} &
        \begin{subfigure}{0.48\linewidth}
            \centering
            \includegraphics[width=\linewidth]{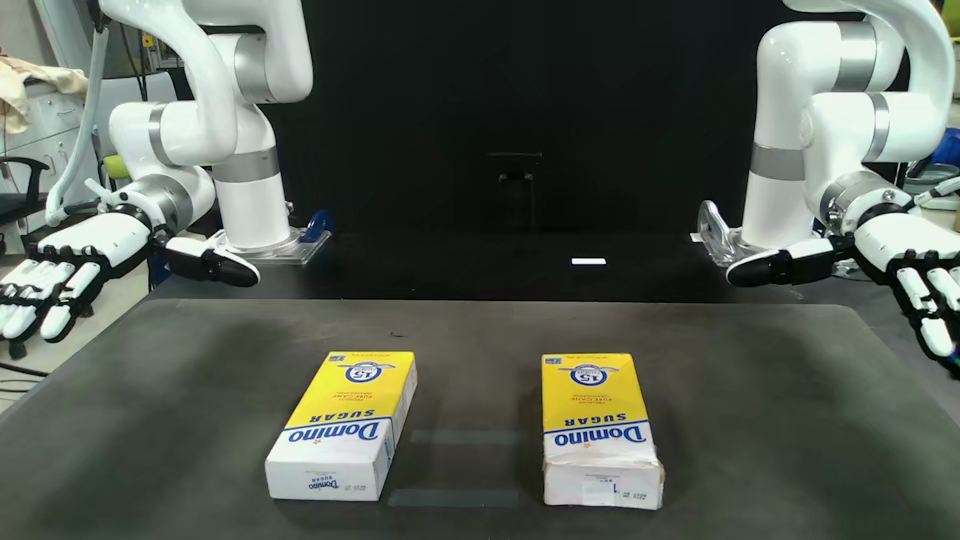}
        \end{subfigure} \\[6pt]

        \begin{subfigure}{0.48\linewidth}
            \centering
            \includegraphics[width=\linewidth]{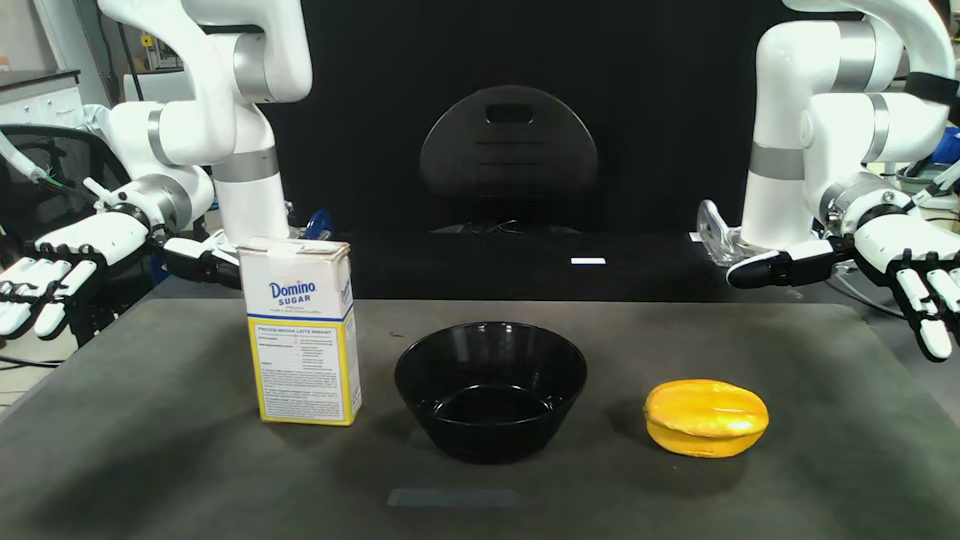}
        \end{subfigure} &
        \begin{subfigure}{0.48\linewidth}
            \centering
            \includegraphics[width=\linewidth]{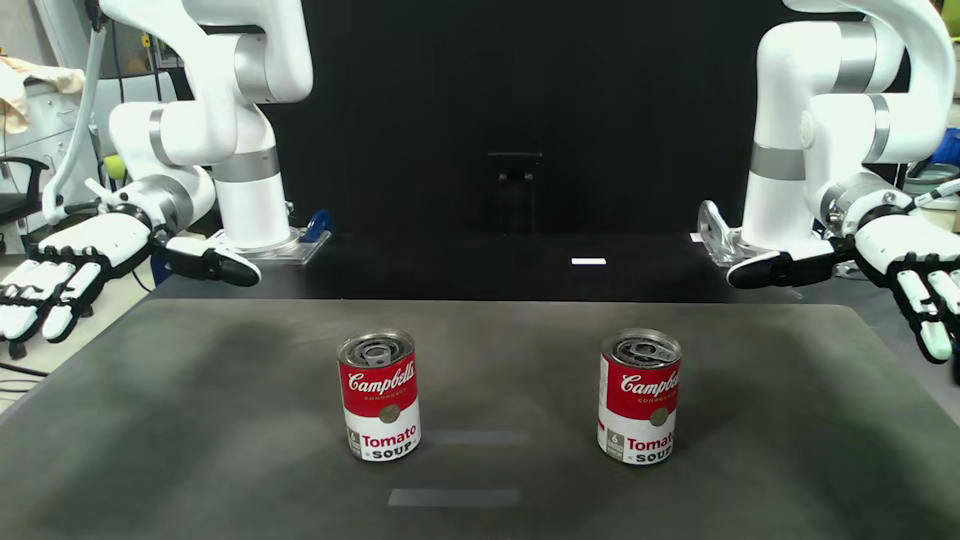}
        \end{subfigure} \\
    \end{tabular}

    \caption{\textbf{Task Visualizations.} We design 10 various tasks to test all the models. The tasks require varied manipulation skills and have different difficulties.}
    \label{fig:task_vis}
\end{figure}

\begin{figure*}[h]
    \centering
    \includegraphics[width=\linewidth]{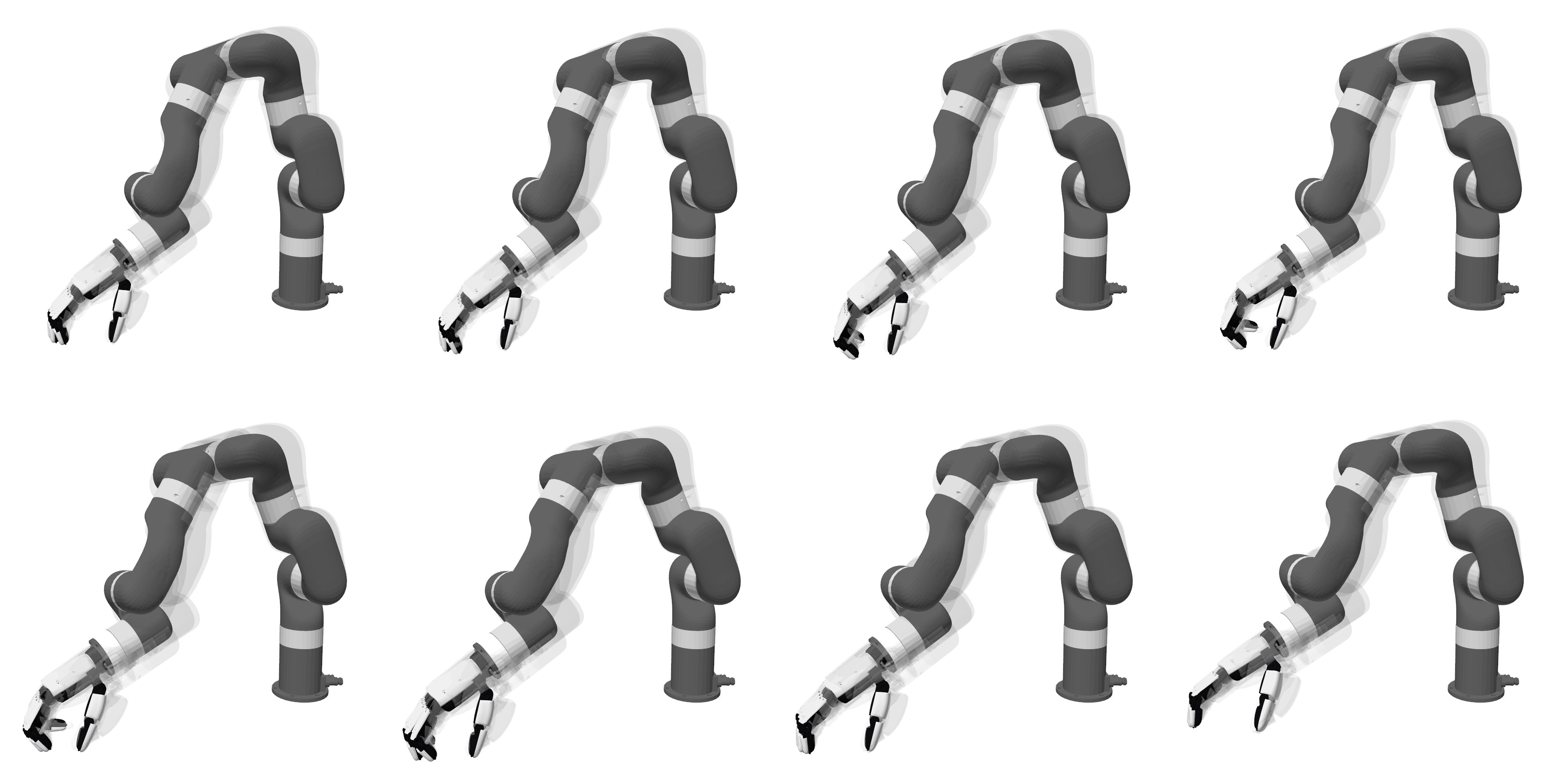}
    \caption{\textbf{Latent Visualization of a Grasping Trajectory.} A trajectory is shown here with all the robot hands.}
    \label{fig:latent_traj}
\end{figure*}

\noindent\textbf{Task Visualizations.}
Using XHand as an example, we show the real-world task visualizations in figure~\ref{fig:task_vis}. Four of the tasks are also tested on G1.

\noindent\textbf{Model Training.}
Besides the model training details provided in the main paper, here are some more training details. RGB images is cropped and then resized from 960 $\times$ 540 to 320 $\times$ 240 during data post-processing. When loaded to train \ourmethod{}, they are resized to 224 $\times$ 224. We use natural language description as the task specification, which is part of the policy condition. The training usually takes around 10 hours for one multi-task policy.

\noindent\textbf{Policy Evaluation.}
For the real-world evaluation, we do 10 trials for each experiment setting. Among these trials, object positions are randomly initialized while the initial joint position for the robot arm and hand remains the same for the same hand.

For unseen tasks only, we record the partial success rate (PSR). If any of the bimanual robot arm finishes its task and the whole task is failed, the overall success rate is 0.5. For other experiments, we do not use PSR. Only rollout that completes a specified task is count as a success.

\subsection{G1 Experiment Results}
Table~\ref{tab:g1-results} is the numeric results for figure~\ref{fig:g1-bar}.
\begin{table}[h]
\centering
\normalsize % Increased from footnotesize
\renewcommand{\arraystretch}{1.1} % Increased row height for "breathing room"
\begin{tabular}{lccccc}
\toprule
\textbf{Method} & \textbf{PF} & \textbf{HB} & \textbf{PS} & \textbf{PoS} & \textbf{Mean} \\
\midrule
\textbf{$\pi_0$~\cite{pi0}}  
& 0.4 
& 0.6  
& 0.5  
& 0.6  
& 0.525 \\

\rowcolor{rowhl}
\textbf{XL-VLA} 
& \textbf{0.7} 
& \textbf{0.9}  
& \textbf{0.9}    
& \textbf{0.8} 
& \textbf{0.825} \impr{57\%} \\
\bottomrule
\end{tabular}
\caption{\textbf{G1 Policy Performances.}}
\label{tab:g1-results}
\end{table}

\end{document}